\documentclass[sigconf]{acmart}

\AtBeginDocument{%
  }

\usepackage{enumitem} 
\begin{document}
\renewcommand\footnotetextcopyrightpermission[1]{}
\settopmatter{printacmref=false} 
\title{PIDiff: Image Customization for Personalized Identities with Diffusion Models}


\author{Jinyu Gu}
\affiliation{%
\institution{School of Computer Science and Information Engineering, Hefei University of Technology}
\streetaddress{}
\city{Hefei}
\country{China}
\state{}}
\email{2023170666@mail.hfut.edu.cn}

\author{Haipeng Liu}
\authornote{Haipeng Liu is the corresponding author}
\affiliation{%
\institution{School of Computer Science and Information Engineering, Hefei University of Technology}
\streetaddress{}
\city{Hefei}
\country{China}
\state{}}
\email{hpliu_hfut@hotmail.com}

\author{Meng Wang}
\affiliation{%
\institution{School of Computer Science and Information Engineering, Hefei University of Technology}
\streetaddress{}
\city{Hefei}
\country{China}
\state{}}
\email{eric.mengwang@gmail.com}

\author{Yang Wang}
\affiliation{%
\institution{School of Computer Science and Information Engineering, Hefei University of Technology}
\streetaddress{}
\city{Hefei}
\country{China}
\state{}}
\email{yangwang@hfut.edu.cn }

\renewcommand{\shortauthors}{}

\begin{abstract}

Text-to-image generation for personalized identities aims at incorporating the specific identity into images using a text prompt and  an identity image. Based on the powerful generative capabilities of denoising diffusion probabilistic models (DDPMs), many previous works adopt additional prompts, such as text embeddings and CLIP image embeddings, to represent the identity information, while they fail to disentangle the identity information and background information. This is because they either mix identity information with text information for backgrounds or extract prompts from content with mixed semantics. As a result, the generated images not only lose key identity characteristics but also suffer from significantly reduced diversity. To address this issue, previous works have combined the $\mathcal{W}_{+}$ space from StyleGAN with diffusion models, leveraging this space to provide a more accurate and comprehensive representation of identity features through multi-level feature extraction. However, the entanglement of identity and background information in in-the-wild images during training prevents accurate identity localization, resulting in severe semantic interference between identity and background. In this paper, we aim to answer two major questions: 1) how to extract personalized identity features accurately and integrate them into the image generation process effectively. 2) how to leverage training strategies to improve the accuracy of visual prompt localization. To this end, we propose a novel fine-tuning-based diffusion model for personalized identities text-to-image generation, named \textbf{PIDiff}, which leverages the $\mathcal{W}_{+}$ space and an identity-tailored fine-tuning strategy  to avoid semantic entanglement and achieves accurate feature extraction and localization. Style editing can also be achieved by PIDiff through preserving the characteristics of identity features in the $\mathcal{W}_{+}$ space, which vary from coarse to fine. Through the combination of the proposed cross-attention block and parameter optimization strategy, PIDiff preserves the identity information and maintains the generation capability for in-the-wild images of the pre-trained model during inference. Our experimental results validate the effectiveness of our method in this task.

\end{abstract}

%

\begin{teaserfigure}
	\centering
	\setlength{\abovecaptionskip}{0pt}
	\setlength{\belowcaptionskip}{5pt}
	\includegraphics[width=0.8\textwidth]{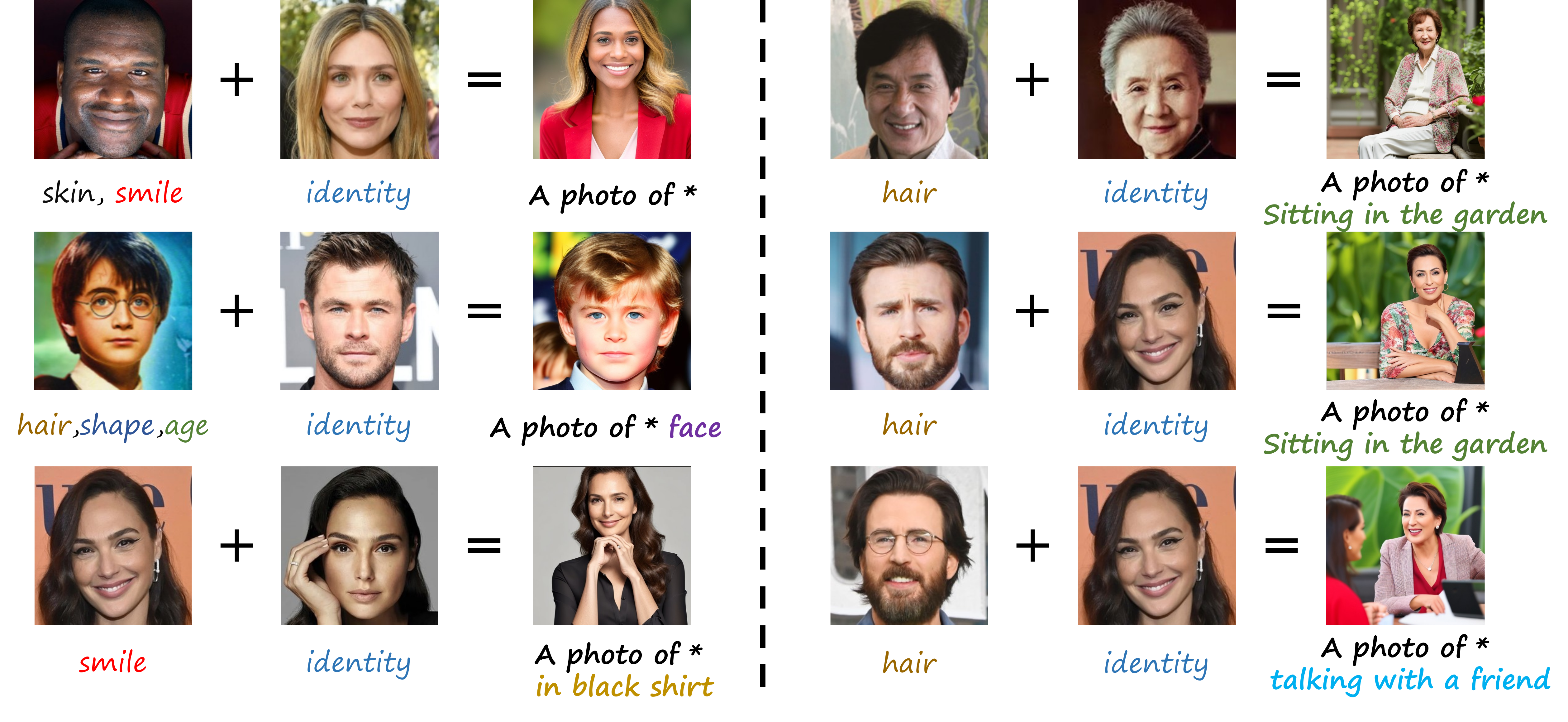}
	\caption{Given face images with coarse-grained features (e.g., hair, skin color, face shape) and fine-grained features (facial details), PIDiff can concatenate the ${w}_{+}$ vectors and generate personalized identity images with new styles}
	\Description{Examples generated by PIDiff using style editing}
	\label{fig:teaser}
		\vspace{-5pt}
\end{teaserfigure}


\maketitle

\section{Introduction}
\label{Introduction}
In recent years, text-to-image generative models \cite{eDiff, GLIDE, DALL, SD, Imagen, RAPHAEL} have attracted significant attention due to their ability to synthesize vivid and diverse images from text prompts. The growing demand for customized content has made text-to-image generation for personalized identities a popular research direction. Specifically, the specific identity is expected to be the main subject of generated images. This introduces two key challenges for generative models: first, how to effectively incorporate the personalized identity into the generated images; second, how to ensure text-image semantic consistency while preserving the unique characteristics of the given identity.

\begin{figure*}[t!]
	\setlength{\abovecaptionskip}{0pt}
	\centering
	\includegraphics[width=0.9\textwidth]{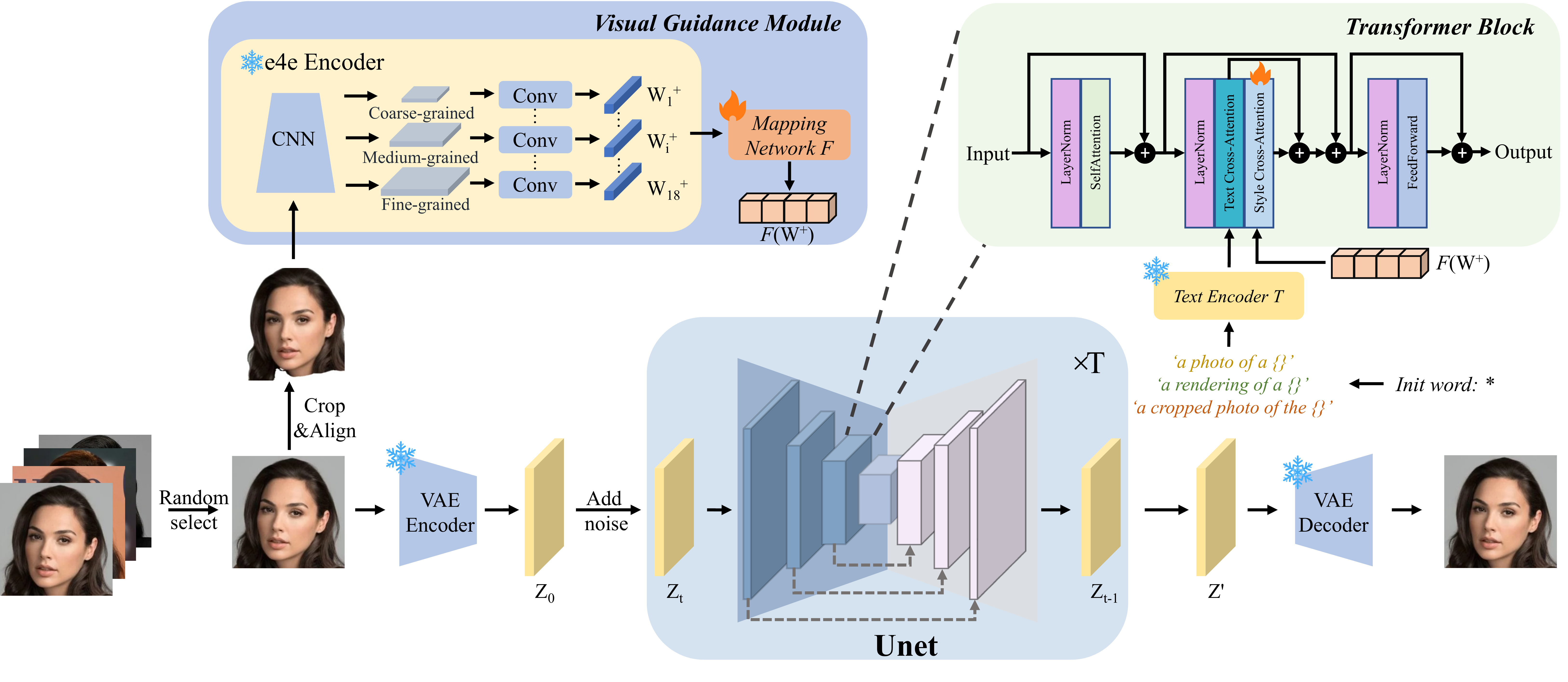}
	
	\caption{Overview of the proposed PIDiff. PIDiff consists of two modules: Visual Guidance Module(VGM) and the diffusion model. VGM processes the preprocessed image and outputs it to the diffusion model. The diffusion model is based on SDV1.5 and uses new transformer blocks to incorporate both text and visual prompts.}
	\label{pipeline}
	\vspace{-12pt}
\end{figure*}	

Recent methods of text-to-image generation have adopted various approaches to represent specific concepts. Some methods \cite{textualinversion, dreambooth, customediff} attempt to inverse specific concepts into the text embedding space. They optimize text embeddings and fine-tune the generative model, allowing it to quickly capture the characteristics of the concept. This strategy allows for the optimization of fewer parameters without compromising the model's performance \cite{hinton2015distilling, wang2024unpacking, qian2023adaptive, qian2023rethinking}. While these methods perform well in generating simple concepts (such as dogs or doors), they face challenges when generating images of personalized identities. As personalized identities often involve many intricate details that require precise representation. Inverting identity into the text embedding space leads to semantic entanglement with textual information, making it difficult to accurately learn and preserve key identity attributes.

To achieve more accurate identity representation, some methods \cite{ip, vico, blipdiff} modify the generative model by adding additional modules to process visual prompts. Although these methods improve feature representation accuracy through additional visual prompts and processing modules, images generated by these methods exhibit poor diversity, and some key features are often overlooked. This is because their approaches to obtain visual prompts is problematic. For example, IP-Adapter \cite{ip} utilizes the image encoder of CLIP. However, extracting visual prompts from image patches is affected by the entanglement of identity and irrelevant region information. As a result, the generated images not only lose crucial personalized identity attributes but also closely resemble the background and some identity attributes of the reference image.

It is worth noting that many works \cite{styleres, w+, hfgi, delving, baykalclip, li2024gradual, pehlivan2023styleres, bobkov2024devil} have utilized the $\mathcal{W}_{+}$ space from StyleGAN for more accurate personalized identity image generation. This is because the encoder \cite{e4e} for $\mathcal{W}_{+}$ space maps the identity to the space through a progressive process, which can provide key identity features more comprehensively. Therefore, $\mathcal{W}_{+}$ Adapter \cite{w+} combines the $\mathcal{W}_{+}$ space with diffusion models to generate more accurate personalized identity images. $\mathcal{W}_{+}$ Adapter trains the model in two stages, where in the second stage, aligned face images are used to reconstruct in-the-wild images. However, in in-the-wild images, a large amount of information from identity-irrelevant regions entangles with the information of identity, making it difficult to accurately localize the visual prompt. As a result, the visual prompts in $\mathcal{W}_{+}$ Adapter not only fail to accurately localize the face region but also severely interfere with the background.

After analyzing the issues with previous methods, we identify two key problems that must be addressed: 1.\textit{how to accurately extract personalized identity features as visual prompts and integrate them into the image generation process} ; 2. \textit{how to train the model to improve the accuracy of visual prompt localization to ensure the preservation of personalized identity features}.

To address the above problems, we propose a novel fine-tuning-based diffusion model called PIDiff for personalized identity text-to-image generation. Due to the excellent performance of diffusion models \cite{eDiff, GLIDE, DALL, SD, Imagen, RAPHAEL, diffusion1, liu2024structure}, we adopt the Stable Diffusion as the generative model. We design a Visual Guidance Module(VGM) to process the reference image and provide the additional visual prompt to the diffusion model. VGM utilize the $\mathcal{W}_{+}$ space of StyleGAN to represent the specific identity. Notably, PIDiff enables personalized identity style editing by preserving the characteristics of the ${w}_{+}$ vector, making the visual prompt more interpretable. During the denoising process, PIDiff utilizes the Style Cross-Attention(SCA) to integrate visual prompts into the image generation process. To improve the accuracy of visual prompt localization and avoid excessive parameter adjustments, we adopt a customized fine-tuning strategy. Our pipeline is illustrated in Fig. \ref{pipeline}.

Our contributions can be summarized as follows:

\begin{enumerate}
	\vspace{-5pt}
	\item  The utilization of the $\mathcal{W}_{+}$ space enables more accurate and comprehensive extraction of personalized identity features. SCA cleverly integrates visual prompts provided by VGM into the image generation process. With a customized fine-tuning strategy, PIDiff avoids semantic entanglement and effectively preserves personalized identity features.

	\item  VGM enables style editing of identities in customized fine-tuning-based methods by preserving the characteristics of the ${w}_{+}$ vector. We are no longer limited to a specific style—PIDiff introduces greater diversity to personalized identity text-to-image generation by allowing style combinations.

	\item To address the limitation of existing datasets that they do not provide multiple high-quality face images for each identity, we propose a small-scale dataset specifically designed for identity image generation. Through extensive experiments, both qualitative and quantitative analyses demonstrate that PIDiff outperforms state-of-the-art methods in personalized identity text-to-image generation ( Our code can be accessed from supplementary material ) .

\end{enumerate}

\section{Methodology}
\label{Sec2.2}

Our work can be summarized into two parts: 1: Customized text-to-image generation for personalized identities:(1) A training strategy for personalized identities (Sec. \ref{2.2.1} and Sec. \ref{2.2.4}). (2) Utilizing the ${w}_{+}$ vector to preserve identity features and enable style editing (Sec. \ref{2.2.2}).  (3) Improving prompt processing capability and training efficiency with a novel Cross-Attention structure (Sec. \ref{2.2.3}). (4) The inference phase (Sec. \ref{2.2.5}). 2: A comprehensive dataset tailored for personalized identity customization (Sec. \ref{2.3}). Before introducing PIDiff, we first elaborate on the preliminaries of diffusion models, which are fundamental to our method.

\subsection{Preliminaries}

\subsubsection{Stable Diffusion}
Stable Diffusion (SD) is a variant of diffusion models, referred to as a latent diffusion model (LDM)~\cite{SD}. It consists of three main components: a Variational Autoencoder (VAE) with an encoder \(E\) and a decoder \(D\), a U-Net~\cite{unet} \( \epsilon_\theta \), and a text encoder~\cite{clip} \( \tau \). It operates by transforming the input image  \(I \in \mathbb{R}^{3 \times H \times W}\) to the latent code \(z_0 \in \mathbb{R}^{4 \times H/8 \times W/8}\), which is in the higher-dimensional latent space, through the VAE encoder \(E\). DDPM~\cite{ddpm} is employed in the training phase for both the forward diffusion process and the reverse denoising process. In the inference phase, DDIM~\cite{ddim} is utilized for the denoising process. Finally, the VAE decoder \(D\) will decode the denoised output back into the pixel space. In this way, diffusion models can not only represent images in an efficient way, but also greatly improve computational efficiency.

A crucial component of SD is the attention mechanism, which consists of both self-attention and cross-attention. The self-attention mechanism allows the model to focus on different parts of the image internally, capturing global dependencies \cite{liu2022delving}. The cross-attention mechanism integrates text conditions into the image generation process, aligning the generated image with the text prompt \(c\). Therefore, inspired by the success of prior methods~  \cite{vico, customediff, w+}, we primarily focus on optimizing cross-attention. We first obtain the latent code \( z \) by adding noise to \( z_0 \) through DDPM, then the cross-attention blocks get query features \( f(z_{t}) \) from the hidden state of input image and text embeddings \( \tau(c) \) from the text encoder \( \tau \). The output of the cross-attention block in the \(i\)-th layer can be defined as:

\begin{equation}
	\label{textCA}
	\begin{split}
		f_{text}^{i'}(z_t) &= \operatorname{Cross-Attention}(Q^{i}, K^{i}, V^{i})  \\
		&= \text{softmax} \left( \frac{Q^{i} (K^{i})^T}{\sqrt{d}} \right) V^{i},
	\end{split}
\end{equation}

where \( Q^{i} = f^{i}(z_t)W^{i}_q \), \( K^{i} = \tau(c)W^{i}_k \), and \( V^{i} = \tau(c)W^{i}_v \) are the query, key, and value matrices of the \(i\)-th cross-attention block, respectively. Specifically, \( W^{i}_q \in \mathbb{R}^{H^{hs} \times W^{hs}} \), \( W^{i}_k \in \mathbb{R}^{H^{hs} \times W^{cd}} \), and \( W^{i}_v \in \mathbb{R}^{H^{hs} \times W^{cd}} \) refer to the projection matrices. Here, \(cd\) denotes the cross-attention dimension, which is the dimensionality of the input features used for cross-attention, corresponding to the text embedding size. \(hs\) represents the hidden state size. The dimension \(d\) of the keys serves to scale the result before applying the softmax function. This mechanism enables the model to align the generated image with the text prompt \(c\) by focusing on relevant semantic features.

After introducing the basic principle, components, and the cross-attention mechanism of SD, the training objective of the diffusion model can be written as:

\begin{equation}
	\mathcal{L}_{\text{LDM}} = \mathbb{E}_{z_0, \epsilon \sim \mathcal{N}(0,1), t, c} \left[ \left\| \epsilon - \epsilon_\theta(z_t, t, \tau(c)) \right\|_2^2 \right],
	\label{eq:ldm_loss}
\end{equation}

where \( z_0 \) is \( E(I) \), \( z_t \) is the latent code at timestep \( t \). \( \epsilon \) is the ground truth noise randomly sampled from a Gaussian distribution. \( t \) is uniformly sampled from \( \{1, 2, \ldots, T\} \). The \( \epsilon_\theta \) represents the U-Net~\cite{unet} denoising network.

\subsubsection{$\mathcal{W}_{+}$ latent space}

Recently, works based on StyleGAN~\cite{StyleGAN} have achieved great success in the task of human face image generation. The $\mathcal{W}_{+}$ latent space possesses several unique characteristics that make it particularly powerful for image generation and manipulation. First, $\mathcal{W}_{+}$ space is multi-dimensional, allowing for fine-grained control over various aspects of the generated image. Additionally, it facilitates style mixing and manipulation by allowing different dimensions of  ${w}_{+}$ vectors to be combined. This flexibility makes $\mathcal{W}_{+}$ latent space an ideal tool for applications such as face image editing, style transfer, and customized image generation.

\subsection{Method}

\begin{figure}[!t]
		\setlength{\abovecaptionskip}{3pt}
	\includegraphics[width=0.42\textwidth ,height=9.2cm,trim=35 0 0 0, clip]{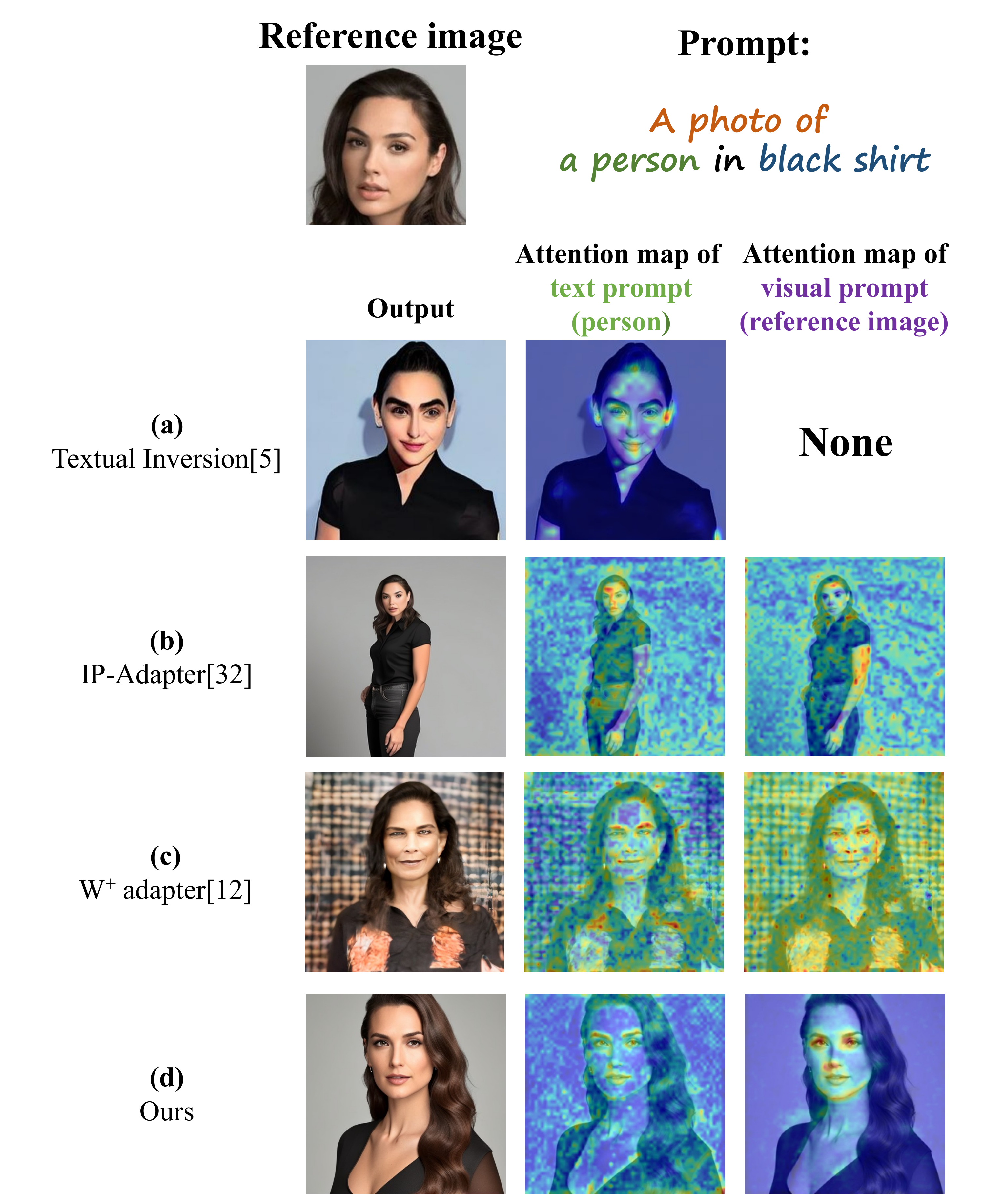}
	\caption{Comparison of Image Generation Results and Attention Maps between Various Methods. The customized training strategy avoids semantic entanglement and effectively preserves key identity features.}
	\label{compare}
	\vspace{-17pt}
\end{figure}

To investigate the reasons behind the lower image quality produced by methods (e.g., IP-Adapter \cite{ip}, $\mathcal{W}_{+}$ Adapter \cite{w+}, and Textual Inversion \cite{textualinversion}), we visualize the attention maps in Fig. \ref{compare}. The following analysis provides insights into the causes of the issues with each method:

\begin{enumerate}[leftmargin=10pt,label={\arabic*)}]
	
	\item  Although Textual Inversion achieves precise attention localization through its training strategy in Fig. \ref{compare}(a), the inherent semantic entanglement and limited expressive power of the text embedding space result in the loss of key identity features.
	
	\item  The attention maps of the visual and text prompts in IP-Adapter can only roughly localize the person’s region and show no significant differences in Fig. \ref{compare}(b). This suggests that CLIP-I causes semantic entanglement by providing embeddings of image patches. Therefore, the visual prompt in IP-Adapter fails to precisely localize facial regions and effectively guide image generation.
	
	\item  In Fig. \ref{compare}(c), the text prompt of $\mathcal{W}_{+}$ Adapter can localize precisely, whereas the visual prompt's attention is dispersed. This is because the semantic entanglement between identity information and background information from in-the-wild images prevents the accurate localization of visual prompts.

\end{enumerate}

Therefore, we employ a customized text-to-image generation strategy to ensure the accurate localization of visual and text prompts. Then, we utilize ${w}_{+}$ vectors as our visual prompts. To preserve the characteristics of the $\mathcal{W}_{+}$ space, we process ${w}_{+}$ vectors through our Visual Guidance Module (VGM). The VGM decomposes the visual prompts into different vectors and feeds them into SD. Finally, through our Style Cross-Attention (SCA), SD can effectively integrate these prompts into the generated images. Our framework is shown in Fig. \ref{pipeline}.

\subsubsection{Customized Text-to-Image Generation}
\label{2.2.1}
Due to the use of pre-trained models, images generated by the model are influenced by prior knowledge. For example, the word ``person" may correspond to the human face images that appear more frequently in the training set. To prevent the generated images of specific identities from being influenced by prior knowledge, we use pseudo-words to represent specific identities, such as \(S^*\).

As shown in Fig.~\ref{pipeline}, during the training process, we only need to provide a few face images of a specific identity, allowing the model to quickly learn accurate localization of visual prompts by avoiding semantic entanglement between identity-relevant and other regions. We randomly select images from provided images as the input. These selected images will be encoded by the VAE encoder and perturbed with random noise according to a randomly selected timestep. In the denoising process, we employ random templates as text prompts, for example, ``an image of \(S^*\) '', ``a cropped photo of \(S^*\) '' and so on. Finally, the UNet outputs the predicted noise based on the text prompt, visual prompt, and the timestep. We will only conduct fine-tuning of the pre-trained model with only a few hundred steps to avoid excessively influencing the pre-trained model. In this way, we enable the model to quickly learn the key features of a specific identity.

During the inference stage, we only need to provide an image of a specific identity and the text prompt that describes the final image. Our model will generate in-the-wild images that not only contain the details of specific identities but also maintain semantic consistency with the text prompt.

\subsubsection{Visual Guidance Module}
\label{2.2.2}
In the task of text-guided image generation for specific identities, it is necessary to ensure the preservation of characteristics of specific identities and semantic consistency. Many previous methods \cite{fastcomposer, ip}  use CLIP-I as the image encoder. However, simply providing image patch embeddings results in the loss of key features due to semantic entanglement, and this also causes a decrease in image diversity. Therefore, we choose the more accurate and flexible \( w_{+} \) vector as our visual prompt. Although the $\mathcal{W}_{+}$ space provides an effective representation for identity, it cannot be directly utilized by SD. Therefore, we design the visual guidance module. First, the preprocessing module can align face images and remove backgrounds to avoid the interference of background. The processed images $\textit{I}_{\text{crop}}$ will be input into the e4e encoder \cite{e4e}. As shown in Fig.~\ref{pipeline}, the e4e encoder first extracts features of the input image from coarse to fine through a CNN, and then obtains the ${w}_{+} \in \mathbb{R}^{18 \times 512}$ vector by mapping modules. Through such multi-level feature extraction, features of the specific identity can be fully extracted. However, the ${w}_{+}$ vector is designed for StyleGAN's generator, so we use a mapping network $\textit{F}$ to project the ${w}_{+}$ vector. Through the mapping network, the ${w}_{+}$ vector can be mapped to a visual embedding $\textit{F}({w}_{+}) \in \mathbb{R}^{4 \times 768}$ to guide the denoising process of the U-Net. The mapping network consists of four mapping layers. Each layer takes part of the ${w}_{+}$ vector as input and outputs a token of dimension 768. The first layer takes the 1-5th latent codes as input, the second layer takes the 6-9th latent codes, the third layer takes the 10-13th latent codes, and the fourth layer takes the 14-18th latent codes. The output of the mapping network is a concatenation of the outputs from these mapping layers.

The visual guidance module serves two important functions. First, it enables the \(w_{+}\) vectors to be utilized by SD. Second, the visual guidance module preserves the properties of the $\mathcal{W}_{+}$ space, making our visual prompts interpretable. As shown in Fig.~\ref{pipeline}, a CNN in the visual prompt module encodes the image into hidden codes of different levels, enabling the final output \(w_{+}\) vector to have multi-level semantic expression capabilities. Consequently, our method enables attribute editing for specific identities, which previous methods \cite{vico, dreambooth, fastcomposer} for customizing specific concepts could not achieve. Specifically, during the training phase, we mainly provide several images of a personalized identity. If these images do not contain the desired style, we can also add an image that includes the additional style we want. Although this may affect the quality of the generated images, our experiments show that our model still produces excellent results. During the inference phase, we can concatenate \(w_{+}\) vectors of images with different styles to achieve style editing to some extent. For example, in Fig.\ref{fig:teaser}, the first example shows that we can use the 1-9th hidden codes of a black man to represent coarse-grained styles of a specific identity, such as skin, age, and the 10-18th hidden codes of a woman to represent fine-grained styles, such as eyes, mouth, and nose. Finally, by concatenating their hidden codes, we can generate a person with desired styles.

\begin{figure}[t!]
	\setlength{\abovecaptionskip}{3pt}
	\centering
	\includegraphics[width=0.94\linewidth]{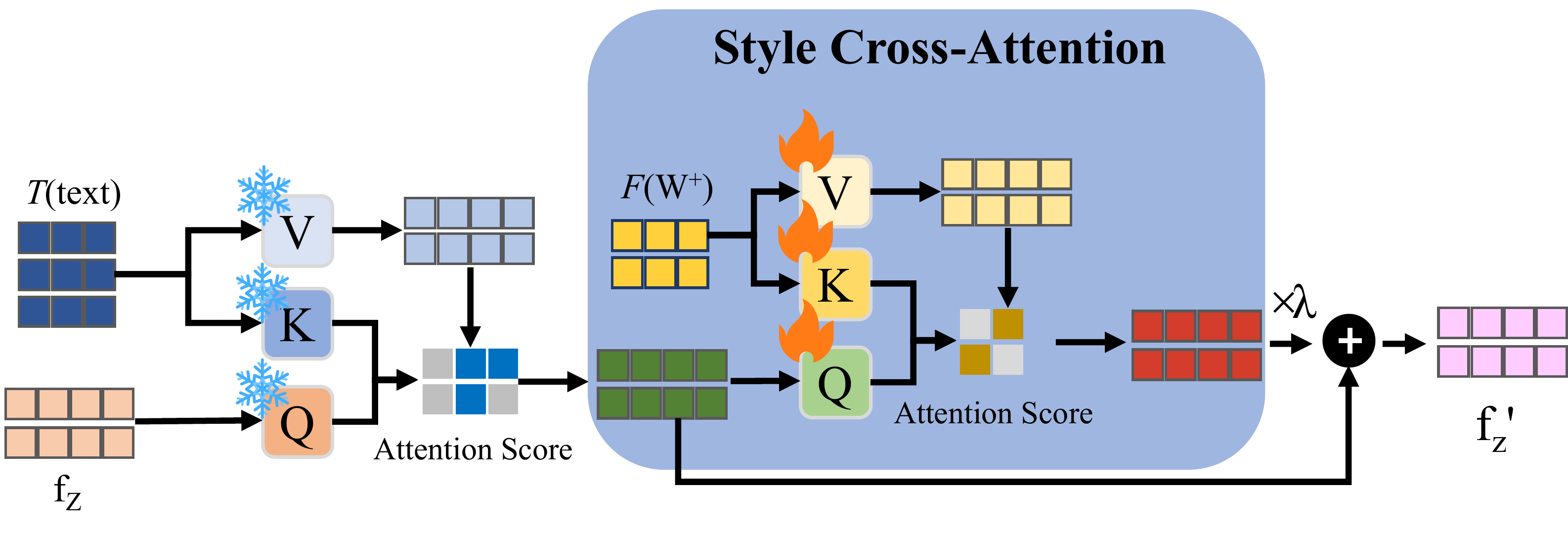}
	\caption{Illustration of Style Cross-Attention(SCA). SCA takes the output of the text cross-attention block as the query and uses the visual prompts as the keys and values. Only the projection matrices is trainable in SCAs.}
	\label{SCA}
	\vspace{-10pt}
\end{figure}

\subsubsection{Style Cross-Attention}
\label{2.2.3}
Previous efforts attempted to achieve personalized identity customization through optimizing text embeddings or fine-tuning diffusion models. By analyzing weight changes after training, Custom Diffusion~\cite{customediff} discovered that, although the cross-attention blocks have relatively few parameters, they have a significant impact on the model's performance. Motivated by these findings, we introduce Style Cross-Attention (SCA) to integrate visual prompts into the denoising process.

As shown in Fig.~\ref{SCA}, SCA is a cross-attention block behind the text cross-attention block. The text cross-attention block is the original cross-attention block for text in the diffusion model and the latent noise interacts with text embeddings in this block. We add SCA for processing visual prompts after the text cross-attention block to integrate visual prompts using semantically richer queries. This structure helps the model localize visual prompts more accurately and effectively prevents the disruption of the text-image semantic consistency of the pre-trained model. Specifically, SCA takes the output of the text cross-attention block as the query and visual embeddings from Mapping Network \(\textit{F}\) as key and value. The output of SCA can be defined as:

\begin{equation}
	\begin{split}
		f_{SCA}^{i''}(z_t) &= \operatorname{Cross-Attention} \left( Q^{i'}, K^{i'}, V^{i'} \right) \\
		&= \text{softmax} \left( \frac{Q^{i'} (K^{i'})^T}{\sqrt{d}} \right) V^{i'},
	\end{split}
\end{equation}

where
\( Q^{i'} = f_{text}^{i'}(z_t)W^{i'}_q \),
\( K^{i'} = F(w_{+})W^{i'}_k \), and
\( V^{i'} = F(w_{+})W^{i'}_v \),
where \( W^{i'}_q \), \( W^{i'}_k \), and \( W^{i'}_v \) are the projection matrices for query, key, and value. \( f_{text}^{i'}(z_t) \) is defined in Eq.~\eqref{textCA}.

Finally, the output of SCA combines the output of text cross-attention block. The final output can be defined as:

\begin{equation}
	f(z_t) = 	f_{text}^{i'}(z_t) + \lambda \cdot f_{SCA}^{i''}(z_t),
\end{equation}

where \(\lambda\) is a parameter that controls the contribution of SCA, which processes visual prompts. During training, \(\lambda\) is set to 1. In the inference stage, \(\lambda\) can be adjusted to balance the semantic information from the text with the visual style derived from the visual prompt.

\subsubsection{Training Loss}
\label{2.2.4}
To accelerate model convergence, \( W^{i'}_q \), \( W^{i'}_k \), \( W^{i'}_v \) are initialized from \( W^{i}_q \), \( W^{i}_k \), \( W^{i}_v \) respectively. During the training process, only \( W^{i'}_q \), \( W^{i'}_k \), \( W^{i'}_v \) in SCAs and the mapping network \(\textit{F}\) will be trainable. This strategy helps PIDiff better preserve the generative capability of the pre-trained model.

The final optimization objective for the model is given as:

\begin{equation}
	\mathcal{L}_{\text{LDM}} = \mathbb{E}_{z_0, \epsilon \sim \mathcal{N}(0,1), t, c} \left[ \left\| \epsilon - \epsilon_\theta(z_t, t, \tau(c), F(w_{+})) \right\|_2^2 \right],
	\label{eq:CSDiff_loss}
\end{equation}

where \eqref{eq:CSDiff_loss} is similar to \eqref{eq:ldm_loss}, except that it incorporates an additional visual condition \( F(w_{+}) \) and adds SCAs after the original text cross-attention blocks in the U-Net.

\begin{figure}[t!]
	\setlength{\abovecaptionskip}{4pt}
	\setlength{\belowcaptionskip}{-10pt}
	\includegraphics[width=0.48\textwidth]{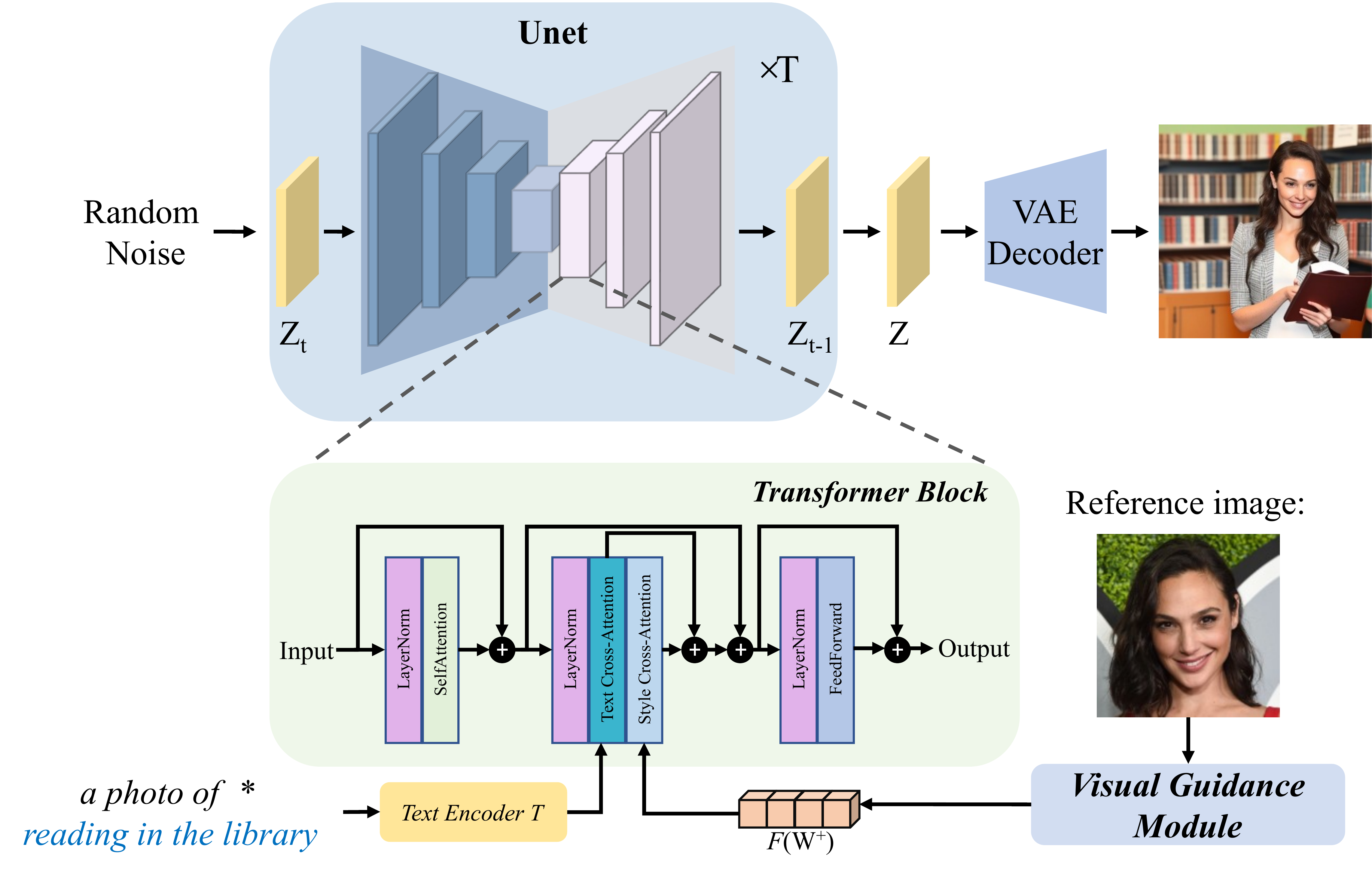}
	\caption{Illustration of inference process in diffusion models: Given a text prompt (e.g.,`` a photo of * reading in the library'') and an identity image, PIDiff can generate images for the identity.}
	\label{Inference}
	\vspace{-5pt}
\end{figure}

\subsubsection{Inference Stage}
\label{2.2.5}
During the inference stage, we utilize Stable Diffusion (SD) as the generative model. As shown in Fig.~\ref{Inference}, this process can be broken down into several key steps. Specifically, we first sample Gaussian noise. This noise serves as the initial latent code for the generation process. Next, we employ the DDIM denoising process to iteratively refine the latent representation. Given a target text prompt, such as ``a photo of * reading in the library '', we obtain the corresponding text embeddings using the text encoder. We also need to provide a specific identity image to guide the generation process through VGM. Finally, the generated image is obtained by decoding the final latent code using the VAE decoder.







\subsection{Dataset Construction Method}
\label{2.3}
\begin{figure}[!t]
	\setlength{\floatsep}{10pt plus 1.0pt minus 2.0pt}
	\setlength{\belowcaptionskip}{-15pt}
	
	\centering
	\includegraphics[width=0.39\textwidth]{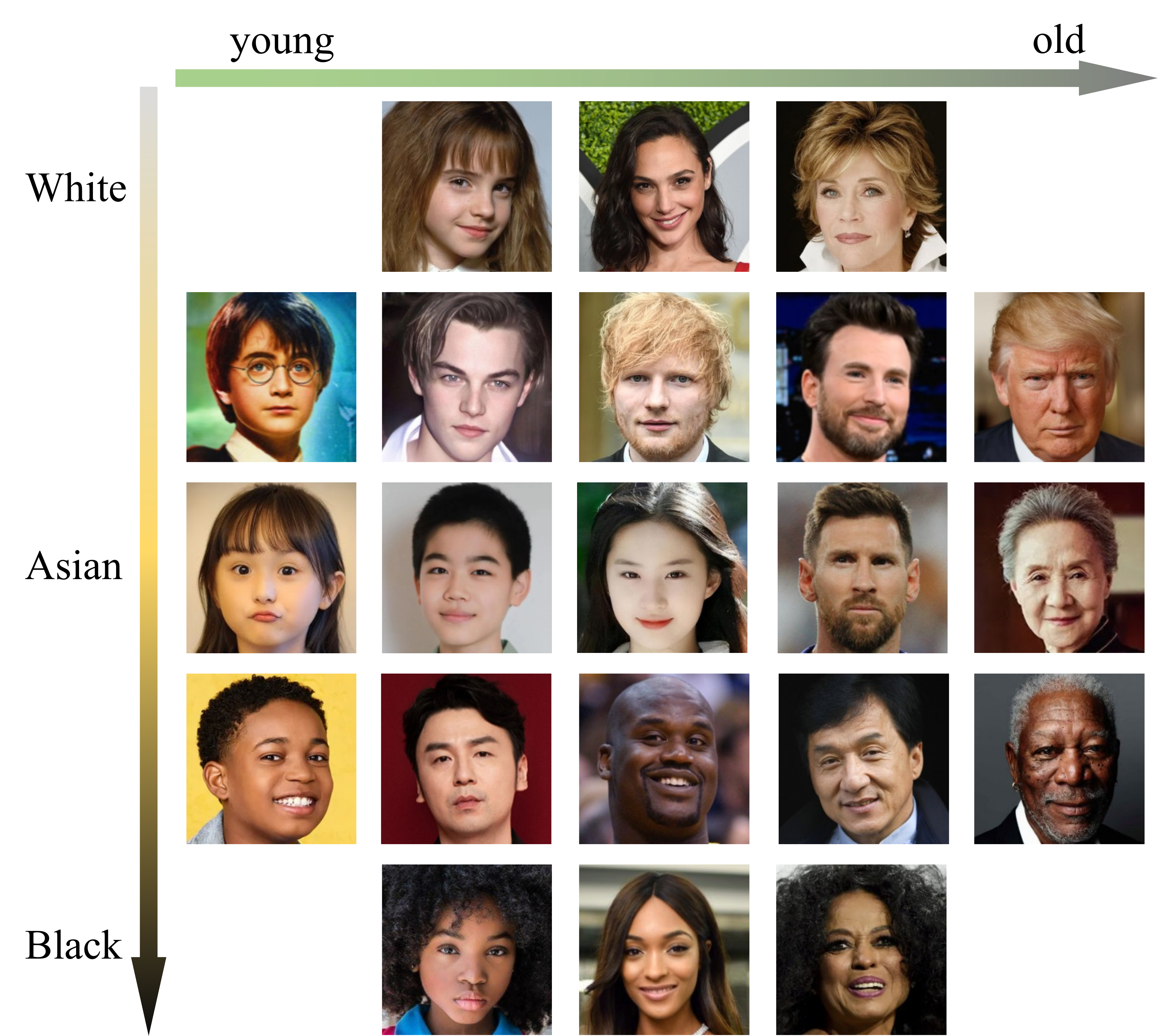}
	\caption{Visualization of our proposed personalized identity customization dataset, we select samples from our dataset showcasing diverse identities, including variations in race, age, and gender.}
	\label{dataset}
	\vspace{-6pt}
\end{figure}

Although existing datasets such as FFHQ contain high-quality face images, they do not offer multiple images for each identity. However, datasets that contain multiple images for each identity are primarily designed for tasks such as image recognition. As a result, these images are often affected by variations in angles and lighting, leading to suboptimal quality. To address this limitation, we constructed a small dataset using images collected from Google for personalized identity customization ( Samples of the dataset are shown in Fig.~\ref{dataset} ) .

Samples from our dataset showcasing diverse identities, including variations in race, age, and gender, for customized text-to-image generation. The dataset consists of 27 identities across three racial groups: White, Asian and Black. Each racial group is further divided into three age brackets: 0–20, 21–50, and 51+, where the age of the samples refers to the age at the time the photo was taken, rather than their current actual age. Both male and female individuals are included within each category, ensuring balanced gender representation.

This carefully curated dataset aims to mitigate potential biases arising from imbalanced representation of demographic groups. Each identity is represented by ten facial images, reflecting variations in expressions and environmental factors. This diversity enhances the dataset’s effectiveness in training and evaluating machine learning models, ensuring more robust experimental results.

By incorporating individuals across various racial, age, and gender groups, our dataset promotes more equitable and accurate research outcomes, fostering a deeper understanding of the complexities in facial recognition technology ( For further analysis, we provide the dataset in the supplementary material ) .

\section{Experiment}

\subsection{Implementation Details}
In this work, we use the pre-trained SD V1.5 as the generative model. We train our model on an A40 GPU. During training, we employ the AdamW optimizer \cite{adaw} with a learning rate of \(1 \times 10^{-4}\) and weight decay of 0.01. We train PIDiff with a batch size of 4 for 600 steps. Unlike the data augmentation strategy of IP-Adapter and \(\mathcal{W}_{+}\) adapter, which use a probability of 0.05 to drop visual and text embeddings, we use a probability of 0.5 ( More analysis can be seen in supplementary material ) . This is because the task scenarios are different. We need the model to quickly learn the features of a specific identity. Therefore, it is necessary not only for the visual prompts to provide features but also for the model itself to learn quickly. We also add random noise to \(w_{+}\) vectors. We adopt DDIM \cite{ddim} with 50 steps during inference. We use the default settings and set the guidance scale to 7.5 to enable classifier-free guidance ( Our code can be accessed from supplementary material ) .

\begin{figure*}[t!]
	\setlength{\abovecaptionskip}{0pt}
	\centering
	\includegraphics[width=0.73\textwidth]{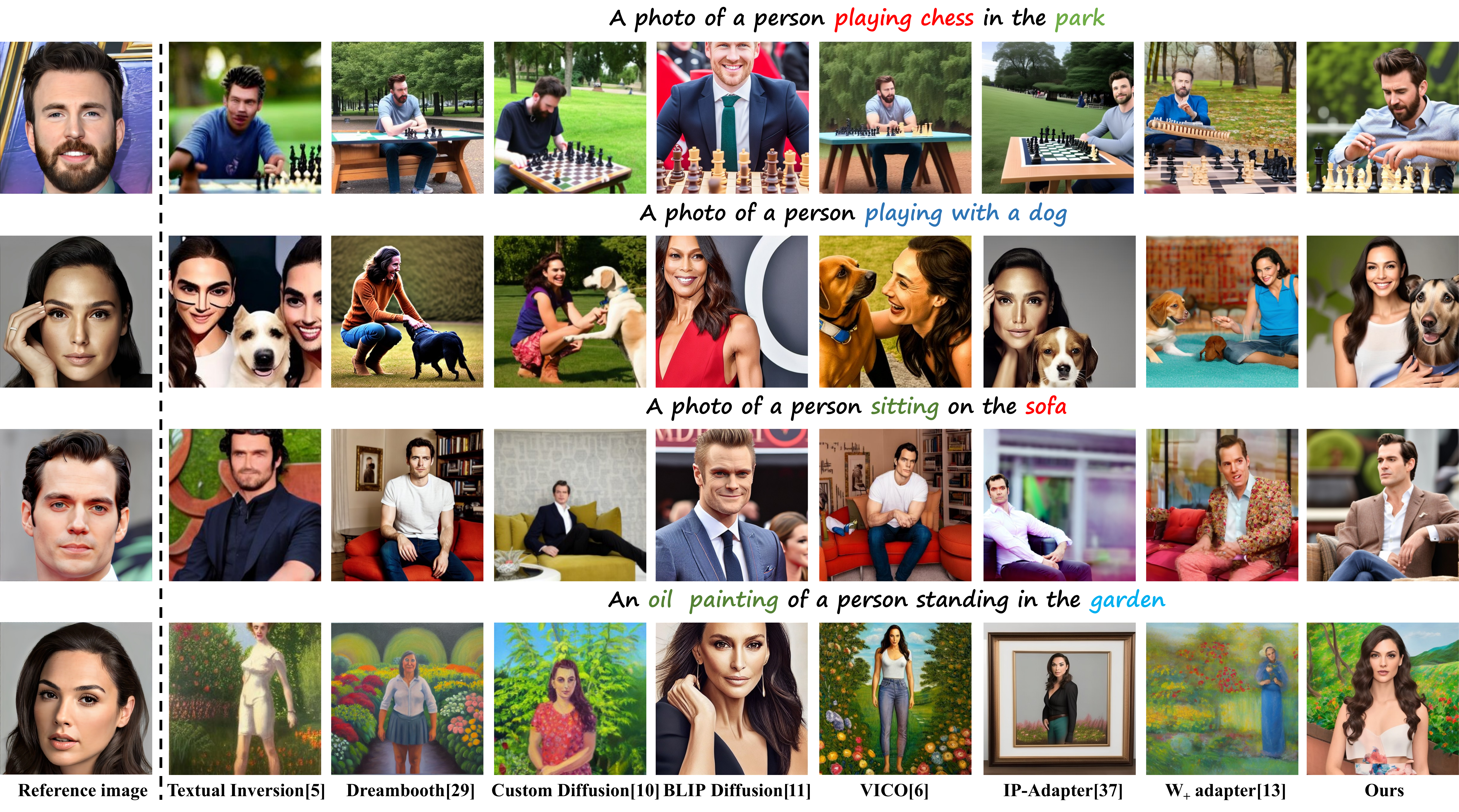}
	\caption{Qualitative Comparison between previous methods and PIDiff. PIDiff not only maintains identity features and text-image semantic consistency but also generates images with significantly high quality.}
	\label{fig5}
\end{figure*}

\subsection{Evaluation Metrics}
We use ID, LPIPS, and CLIP-T to evaluate the performance of our model. \textbf{Identity Loss (ID$\uparrow$)}: We first use MTCNN \cite{mtcnn} for face alignment to prevent measurement errors. Then we use ArcFace \cite{arcface} to measure detected faces. Finally, we assess the identity preservation by calculating the cosine similarity between the feature of the generated image and the original image. \textbf{Learned Perceptual Image Patch Similarity (LPIPS$\downarrow$)} \cite{lpips}: We use VGG-V0.1 for image feature extraction and evaluate image similarity by comparing the differences between features, where the differences are assessed directly by calculating the squared differences. \textbf{Text-image similarity (CLIP-T$\uparrow$)} \cite{clip-T}: We use the pre-trained clip-vit-base-patch16 to calculate the similarity between the generated image and the text prompt.

\begin{figure}[!t]
	\setlength{\abovecaptionskip}{0pt}
	\centering
	\includegraphics[width=0.4\textwidth]{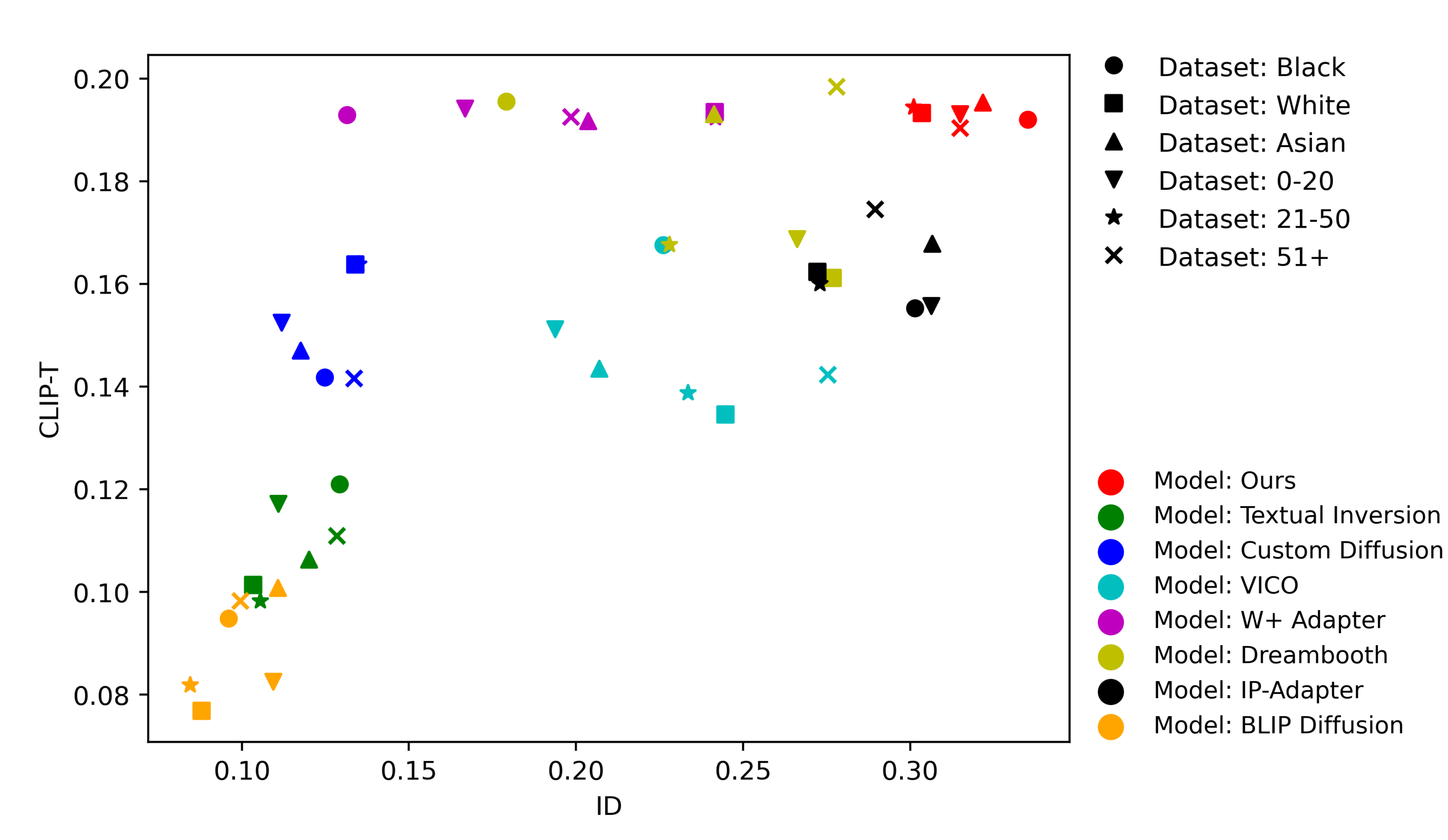}
	\caption{Comparative Analysis of Models Across Various Datasets. Compared to other methods, our approach is free from bias. Outstanding experimental results demonstrate that our method is both superior and more stable.}
	\label{pointres}
	\vspace{-16pt}
\end{figure}

\begin{figure}[!t]
		\vspace{3pt}
	\setlength{\abovecaptionskip}{0pt}
	\centering
	\includegraphics[width=0.45\textwidth]{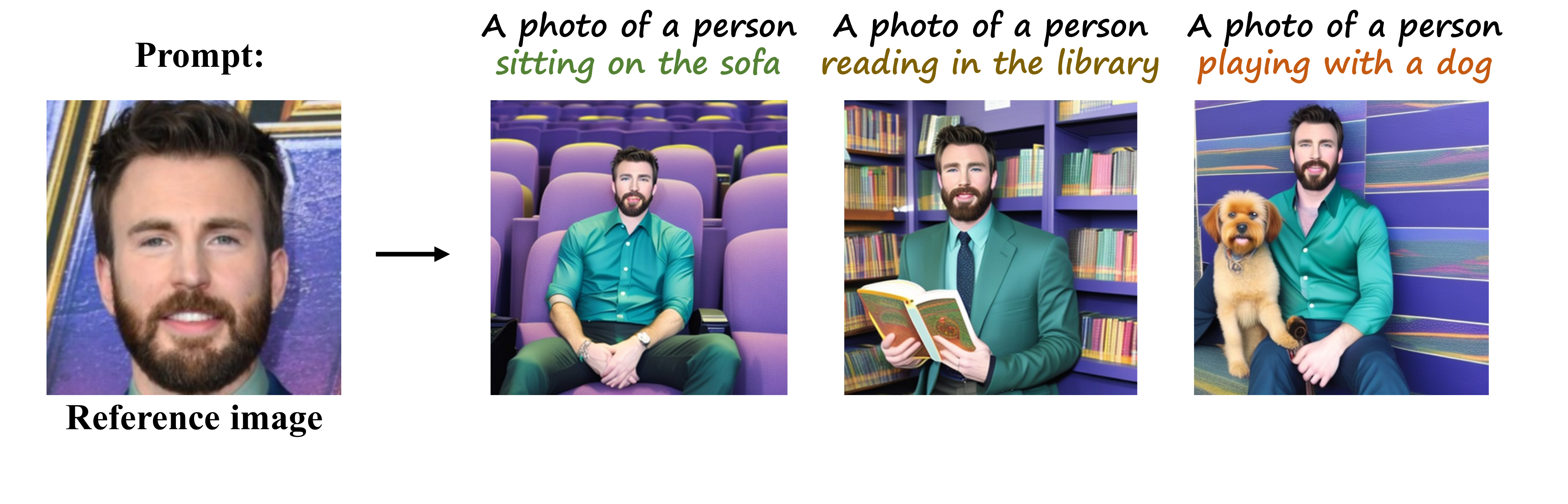}
	\caption{Examples generated by the model \cite{ip} using CLIP-I as the image encoder. The facial features are overly fixed, and the backgrounds of these images are severely affected.}
	\label{clipimages}
	\vspace{-18pt}
\end{figure}

\subsection{Comparison with State-of-the-arts}

To validate the superiority of PIDiff, we compare it with typical models, including: Textual Inversion \cite{textualinversion} proposes finding text embeddings for different concepts. Custom Diffusion \cite{customediff} introduces fine-tuning the cross-attention mapping matrix of diffusion models and text embeddings. DreamBooth \cite{dreambooth} attempts to represent specific concepts using unique identifiers. BLIP-Diffusion \cite{blipdiff} also attempts to generate images in the text embedding space. We demonstrate through experimental data that relying solely on text embedding space cannot effectively preserve personalized identity features. VICO \cite{vico} and IP-Adapter \cite{ip} attempt to use additional modules to integrate visual prompts, but the image encoders they use not only fail to accurately capture key features but also reduce the diversity of the generated images. $\mathcal{W}_{+}$ adapter \cite{w+} utilizes the $\mathcal{W}_{+}$ space to achieve high-quality identity representation and enables image generation for arbitrary identities, but its training strategy leads to a decline in the quality of the generated images.

\begin{table}[H]
	\begin{center}
		\vspace{-7pt}
		\caption{ Quantitative comparisons with previous methods. We classified previous methods in the table according to Sec.\ref{Introduction} to further validate the effectiveness of our method. Results highlighted by \textbf{bold} and \underline{underline} represent the first and second best results. }
		\label{tab1}
		\begin{tabular}{c | c  c  c }
			\hline
			Methods & ID$\uparrow$ & LPIPS$\downarrow$ & CLIP-T$\uparrow$\\
			\hline
			Textual Inversion\cite{textualinversion} & 0.1123 & 0.6717 & 0.1063\\
			
			Custom Diffusion\cite{customediff} & 0.1284 & 0.7331 & 0.1557 \\
			
			BLIP Diffusion\cite{blipdiff} & 0.0947 & 0.6159& 0.0859 \\
			
			Dreambooth\cite{dreambooth} & 0.2498 & 0.6778& 0.1751 \\
			
			\hline
			
			VICO\cite{vico}  & 0.2325& 0.6873 & 0.1430\\
			IP-Adapter\cite{ip} & \underline{0.2858} & \underline{0.5947}& 0.1623 \\
			\hline
			
			$\mathcal{W}_{+}$ adapter\cite{w+} & 0.2668 & 0.6774 & \underline{0.1935} \\
			\hline
			Ours  & \textbf{0.3112} & \textbf{0.5936}& \textbf{0.1938} \\
			\hline
			
		\end{tabular}
	\end{center}
	\footnotesize 
	\begin{flushleft}
		
	\end{flushleft}
	\vspace{-15pt}
\end{table}

\subsubsection{Quantitative Comparison}
We compared our model with the state-of-the-art text-to-image generation models. In the experiment, we use 12 text prompts as text conditions for each identity. These text prompts include scenarios with single people, multiple people, and multiple objects, as well as requirements for clothing and poses. Since some models, such as Textual Inversion, Custom Diffusion, do not require reference images during inference, we  select the most similar facial image from the training images to compute the evaluation metrics. This comprehensive evaluation ensures an accurate performance assessment for each model.

As shown in Table~\ref{tab1}, our model achieves outstanding results. Notably, our method attains higher ID scores. This advantage stems from our visual prompt being based on the \(\mathcal{W}_{+}\) space, which provides superior expressive power. Additionally, our approach outperforms $\mathcal{W}_{+}$ adapter. This is because they are affected by the training strategy, which causes visual prompts to fail in accurately localizing relevant regions (e.g., Fig.~\ref{compare}(c)).

Our method also achieves higher CLIP-T metric, benefiting from the fusion of the training strategy and SCA. As stated in Sec.~\ref{2.2.1} and Sec.~\ref{2.2.3}, \textit{the customized training strategy allows the model to quickly learn accurate localization of visual prompts by avoiding semantic entanglement. SCA effectively prevents the disruption of the text-image semantic consistency of the pre-trained model.}

Fig.~\ref{pointres} presents our results across various test categories, including different ethnicities and age groups. The stable results suggest that our model is free from bias and is suitable for text-to-image generation tasks for a wide range of identities.

\begin{figure*}[!t]
	\setlength{\abovecaptionskip}{0pt}
	\setlength{\belowcaptionskip}{0pt}
	\centering
	\vspace{-55pt} 
	\includegraphics[width=0.8\linewidth,height=8.6cm,trim=160 90 190 90, clip]{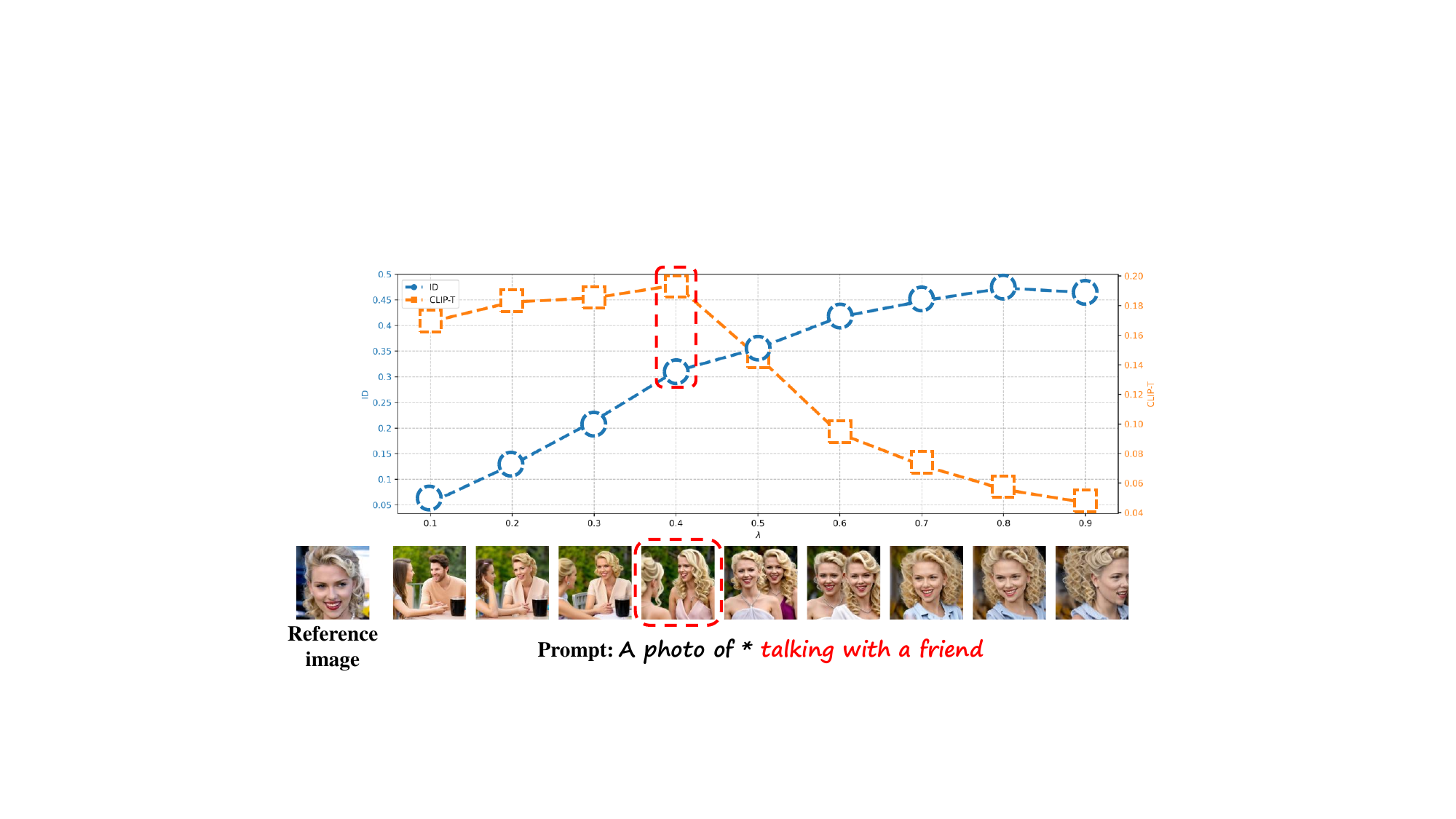}
	\caption{Visual comparisons of images generated by using different \(\lambda\) in SCA. When \(\lambda\)=0.4, PIDiff achieves the highest text-image semantic consistency while effectively preserving identity features.}
	\label{scale}
	\vspace{-10pt}
\end{figure*}

\begin{figure}[!t]
	\setlength{\abovecaptionskip}{0pt}
	\setlength{\belowcaptionskip}{-12pt}
	\centering
	\includegraphics[width=\linewidth]{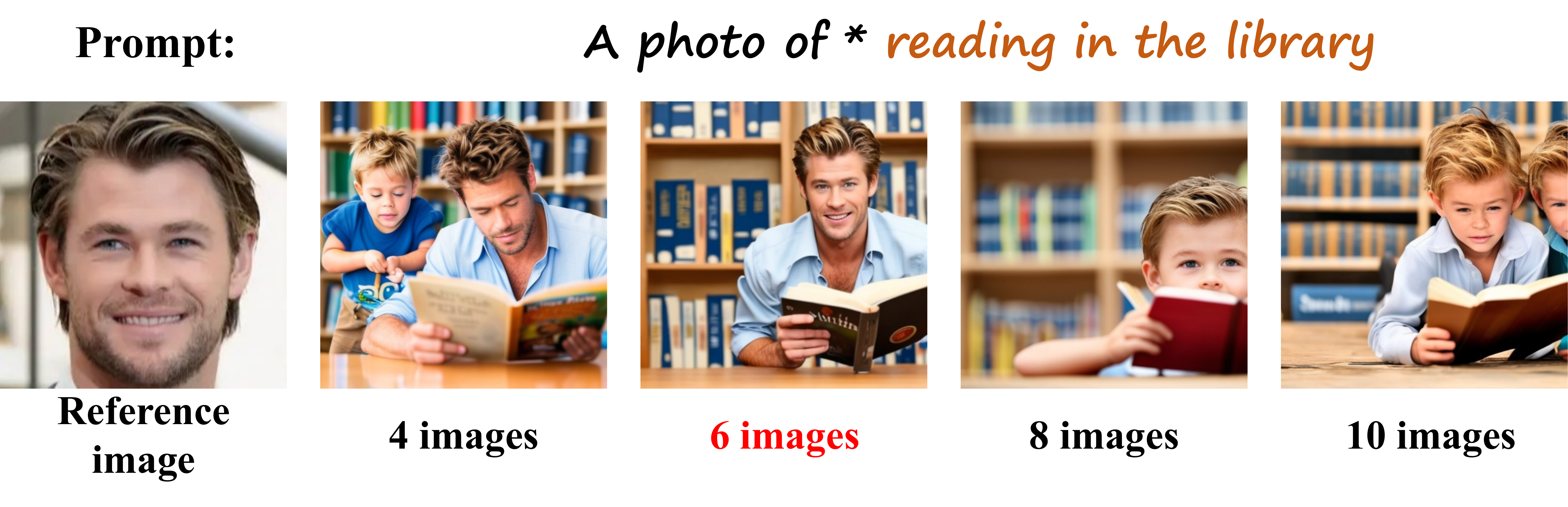}
	\caption{Qualitative results of using different numbers of training images}
	\label{numimages}
		\vspace{-2pt}
\end{figure}

\subsubsection{Qualitative Comparison}

Fig. \ref{fig5} demonstrates the visual comparison between PIDiff and other methods. It can be observed that methods without visual prompts have poor identity preservation capabilities in the generated images. While identity preservation is essential, it is also important for the pose and background to vary based on text prompts. We can notice that faces in the images generated by IP-Adapter are relatively fixed, and the background is affected by visual prompts(see Fig.~\ref{clipimages} and IP-Adapter rows 2 and 4 in Fig.~\ref{fig5}). This aligns with the issue regarding CLIP image encoder that we highlighted in Sec.~\ref{2.2.2}: \textit{Simply providing image patch embeddings results in the loss of key features due to semantic entanglement, and this also causes a decrease in image diversity.}

These results further validate the effectiveness of the $\mathcal{W}_{+}$ latent space and the Visual Guidance Module in our method. By aligning faces and cropping backgrounds, the visual prompts provided by the preprocessing module effectively mitigate background interference. Through VGM's multi-level processing approach, the visual prompts can be more effectively utilized by SCA.

\subsection{Ablation Study}

\subsubsection{Analysis of Style Editing}

Through the use of ${w}_{+}$ vector, our model is capable of performing a certain degree of style editing. We can synthesize images of identities with specific styles by combining the ${w}_{+}$ vectors of the specific identity. If images of a specific identity do not have the desired style, we can add an image with the specific style to the training set. As demonstrated in Fig. \ref{fig:teaser}, the first image provides the desired style, and the second provides the fine-grained characteristics that represent the specific identity. By fusing \(w_{+}\) vectors, we can add a new style to the specific identity.


\subsubsection{Analysis of Number of Images for Train}

Previous methods typically require a few images to customize specific concepts. However, identity face has many unique characteristics that need to be preserved. Therefore, the number of images used for training must be reasonable. We choose to use six images for training. In Fig. \ref{numimages}, we show the visual comparison between our choice and other numbers, and we also show experiment results in Table~\ref{numtable}. It can be seen that when there are fewer images for training, the model tends to overfit images and text prompts in dataset. In the inference stage, the generated image is easily affected by the text prompt, which leads to the degradation of image quality. When there are too many images, the characteristics of identity are difficult to learn.

\subsubsection{Analysis of Style Cross-Attention Structure }

In IP-Adapter \cite{ip}, a parallel cross-attention block(PCA) design is adopted, where the query for the visual cross-attention block comes directly from the hidden state. However, in SCA, the query originates  from the output of the text cross-attention block. Therefore, we show experimental comparisons in table \ref{numtable}. We found that, due to the processing by the text cross-attention block, different semantic regions in the image are better distinguished, allowing the visual prompt to be more precisely localized. As a result, our SCA can help the visual prompt focus on the facial region more accurately, ensuring the retention of identity features.

\begin{table}
	\begin{center}
		\caption{Quantitative Comparison of Training Configurations and Style Cross-Attention Structure }
		\label{numtable}
		\begin{tabular}{c | c  c  c c }
			\hline
			\multicolumn{5}{c}{Analysis of the Number of Training Images } \\
			\hline
			images & 4 & 6 & 8  & 10\\
			\hline
			ID$\uparrow$ & 0.2457 & \textbf{0.3112} & 0.2888 & 0.2883\\
			
			CLIP-T$\uparrow$ & 0.1410 & \textbf{0.1938} & 0.1663 & 0.1477 \\
			\hline
			
			\multicolumn{5}{c}{Analysis of Style Cross-Attention Structure } \\
			\hline
			Structure  &\multicolumn{2}{c}{PCA} &\multicolumn{2}{c}{SCA}\\
			\hline
			ID $\uparrow$& \multicolumn{2}{c}{0.2235}  & \multicolumn{2}{c}{\textbf{0.3112}}\\
			
			CLIP-T$\uparrow$ & \multicolumn{2}{c}{0.1468}  &\multicolumn{2}{c}{ \textbf{0.1938}} \\
			\hline
			
		\end{tabular}
	\end{center}
	\vspace{-16pt}
\end{table}

\subsubsection{Analysis of \(\lambda\) in Style Cross-Attention}
We use \(\lambda\) to control the influence of \(w_{+}\) vectors on the hidden states in SCA. As shown in the Fig. \ref{scale}, when \(\lambda\) approaches 0, the generated images retain the text alignment capabilities of the pre-trained SD, but the specific identity is not well preserved. When \(\lambda\) approaches 1, the generated image fails to match the text prompt. It can be seen that when \(\lambda\) is smaller than 0.4, the  features of the specific identity are lost. When the \(\lambda\) is larger than 0.4, CLIP-T rapidly decreases, while ID does not change significantly. Therefore, through experimental analysis, we choose 0.4 as an appropriate choice.

\section{Conclusion}

In this paper, we propose PIDiff for personalized identities text-to-image generation, which utilizes the $\mathcal{W}_{+}$ space and diffusion models to achieve personalized identity text-to-image generation. We demonstrating that: 1) The $\mathcal{W}_{+}$ space enhances the diffusion model's accuracy in representing identity features and enables flexible style editing. 2) The cross-attention mechanism and customized fine-tuning training strategy effectively avoid semantic entanglement and improve semantic consistency between text and image prompts. Extensive experimental results validate that PIDiff is free from bias and outperforms previous methods.

\bibliographystyle{ACM-Reference-Format}
\bibliography{sample-base}


\begin{thebibliography}{39}


\ifx \showCODEN    \undefined \def \showCODEN     #1{\unskip}     \fi
\ifx \showDOI      \undefined \def \showDOI       #1{#1}\fi
\ifx \showISBNx    \undefined \def \showISBNx     #1{\unskip}     \fi
\ifx \showISBNxiii \undefined \def \showISBNxiii  #1{\unskip}     \fi
\ifx \showISSN     \undefined \def \showISSN      #1{\unskip}     \fi
\ifx \showLCCN     \undefined \def \showLCCN      #1{\unskip}     \fi
\ifx \shownote     \undefined \def \shownote      #1{#1}          \fi
\ifx \showarticletitle \undefined \def \showarticletitle #1{#1}   \fi
\ifx \showURL      \undefined \def \showURL       {\relax}        \fi
\providecommand\bibfield[2]{#2}
\providecommand\bibinfo[2]{#2}
\providecommand\natexlab[1]{#1}
\providecommand\showeprint[2][]{arXiv:#2}

\bibitem[\protect\citeauthoryear{Balaji, Nah, Huang, Vahdat, Song, Zhang,
  Kreis, Aittala, Aila, Laine, et~al\mbox{.}}{Balaji et~al\mbox{.}}{2022}]%
        {eDiff}
\bibfield{author}{\bibinfo{person}{Yogesh Balaji}, \bibinfo{person}{Seungjun
  Nah}, \bibinfo{person}{Xun Huang}, \bibinfo{person}{Arash Vahdat},
  \bibinfo{person}{Jiaming Song}, \bibinfo{person}{Qinsheng Zhang},
  \bibinfo{person}{Karsten Kreis}, \bibinfo{person}{Miika Aittala},
  \bibinfo{person}{Timo Aila}, \bibinfo{person}{Samuli Laine}, {et~al\mbox{.}}}
  \bibinfo{year}{2022}\natexlab{}.
\newblock \showarticletitle{ediff-i: Text-to-image diffusion models with an
  ensemble of expert denoisers}.
\newblock \bibinfo{journal}{\emph{arXiv preprint arXiv:2211.01324}}
  (\bibinfo{year}{2022}).
\newblock


\bibitem[\protect\citeauthoryear{Baykal, Anees, Ceylan, Erdem, Erdem, and
  Yuret}{Baykal et~al\mbox{.}}{2023}]%
        {baykalclip}
\bibfield{author}{\bibinfo{person}{Ahmet~Canberk Baykal},
  \bibinfo{person}{Abdul~Basit Anees}, \bibinfo{person}{Duygu Ceylan},
  \bibinfo{person}{Erkut Erdem}, \bibinfo{person}{Aykut Erdem}, {and}
  \bibinfo{person}{Deniz Yuret}.} \bibinfo{year}{2023}\natexlab{}.
\newblock \showarticletitle{CLIP-guided StyleGAN Inversion for Text-driven Real
  Image Editing}.
\newblock \bibinfo{journal}{\emph{ACM Transactions on Graphics}}
  \bibinfo{volume}{42}, \bibinfo{number}{5} (\bibinfo{year}{2023}),
  \bibinfo{pages}{1--18}.
\newblock


\bibitem[\protect\citeauthoryear{Bobkov, Titov, Alanov, and Vetrov}{Bobkov
  et~al\mbox{.}}{2024}]%
        {bobkov2024devil}
\bibfield{author}{\bibinfo{person}{Denis Bobkov}, \bibinfo{person}{Vadim
  Titov}, \bibinfo{person}{Aibek Alanov}, {and} \bibinfo{person}{Dmitry
  Vetrov}.} \bibinfo{year}{2024}\natexlab{}.
\newblock \showarticletitle{The devil is in the details: Stylefeatureeditor for
  detail-rich stylegan inversion and high quality image editing}. In
  \bibinfo{booktitle}{\emph{Proceedings of the IEEE/CVF Conference on Computer
  Vision and Pattern Recognition}}. \bibinfo{pages}{9337--9346}.
\newblock


\bibitem[\protect\citeauthoryear{Deng, Guo, Xue, and Zafeiriou}{Deng
  et~al\mbox{.}}{2019}]%
        {arcface}
\bibfield{author}{\bibinfo{person}{Jiankang Deng}, \bibinfo{person}{Jia Guo},
  \bibinfo{person}{Niannan Xue}, {and} \bibinfo{person}{Stefanos Zafeiriou}.}
  \bibinfo{year}{2019}\natexlab{}.
\newblock \showarticletitle{Arcface: Additive angular margin loss for deep face
  recognition}. In \bibinfo{booktitle}{\emph{Proceedings of the IEEE/CVF
  conference on computer vision and pattern recognition}}.
  \bibinfo{pages}{4690--4699}.
\newblock


\bibitem[\protect\citeauthoryear{Gal, Alaluf, Atzmon, Patashnik, Bermano,
  Chechik, and Cohen-Or}{Gal et~al\mbox{.}}{2022}]%
        {textualinversion}
\bibfield{author}{\bibinfo{person}{Rinon Gal}, \bibinfo{person}{Yuval Alaluf},
  \bibinfo{person}{Yuval Atzmon}, \bibinfo{person}{Or Patashnik},
  \bibinfo{person}{Amit~H Bermano}, \bibinfo{person}{Gal Chechik}, {and}
  \bibinfo{person}{Daniel Cohen-Or}.} \bibinfo{year}{2022}\natexlab{}.
\newblock \showarticletitle{An image is worth one word: Personalizing
  text-to-image generation using textual inversion}.
\newblock \bibinfo{journal}{\emph{arXiv preprint arXiv:2208.01618}}
  (\bibinfo{year}{2022}).
\newblock


\bibitem[\protect\citeauthoryear{Hao, Han, Zhao, and Wong}{Hao
  et~al\mbox{.}}{2023}]%
        {vico}
\bibfield{author}{\bibinfo{person}{Shaozhe Hao}, \bibinfo{person}{Kai Han},
  \bibinfo{person}{Shihao Zhao}, {and} \bibinfo{person}{Kwan-Yee~K Wong}.}
  \bibinfo{year}{2023}\natexlab{}.
\newblock \showarticletitle{ViCo: Plug-and-play Visual Condition for
  Personalized Text-to-image Generation}.
\newblock \bibinfo{journal}{\emph{arXiv preprint arXiv:2306.00971}}
  (\bibinfo{year}{2023}).
\newblock


\bibitem[\protect\citeauthoryear{Hinton, Vinyals, and Dean}{Hinton
  et~al\mbox{.}}{2015}]%
        {hinton2015distilling}
\bibfield{author}{\bibinfo{person}{Geoffrey Hinton}, \bibinfo{person}{Oriol
  Vinyals}, {and} \bibinfo{person}{Jeff Dean}.}
  \bibinfo{year}{2015}\natexlab{}.
\newblock \showarticletitle{Distilling the knowledge in a neural network}.
\newblock \bibinfo{journal}{\emph{arXiv preprint arXiv:1503.02531}}
  (\bibinfo{year}{2015}).
\newblock


\bibitem[\protect\citeauthoryear{Ho, Jain, and Abbeel}{Ho
  et~al\mbox{.}}{2020}]%
        {ddpm}
\bibfield{author}{\bibinfo{person}{Jonathan Ho}, \bibinfo{person}{Ajay Jain},
  {and} \bibinfo{person}{Pieter Abbeel}.} \bibinfo{year}{2020}\natexlab{}.
\newblock \showarticletitle{Denoising diffusion probabilistic models}.
\newblock \bibinfo{journal}{\emph{Advances in neural information processing
  systems}}  \bibinfo{volume}{33} (\bibinfo{year}{2020}),
  \bibinfo{pages}{6840--6851}.
\newblock


\bibitem[\protect\citeauthoryear{Karras}{Karras}{2019}]%
        {StyleGAN}
\bibfield{author}{\bibinfo{person}{Tero Karras}.}
  \bibinfo{year}{2019}\natexlab{}.
\newblock \showarticletitle{A Style-Based Generator Architecture for Generative
  Adversarial Networks}.
\newblock \bibinfo{journal}{\emph{arXiv preprint arXiv:1812.04948}}
  (\bibinfo{year}{2019}).
\newblock


\bibitem[\protect\citeauthoryear{Kumari, Zhang, Zhang, Shechtman, and
  Zhu}{Kumari et~al\mbox{.}}{2023}]%
        {customediff}
\bibfield{author}{\bibinfo{person}{Nupur Kumari}, \bibinfo{person}{Bingliang
  Zhang}, \bibinfo{person}{Richard Zhang}, \bibinfo{person}{Eli Shechtman},
  {and} \bibinfo{person}{Jun-Yan Zhu}.} \bibinfo{year}{2023}\natexlab{}.
\newblock \showarticletitle{Multi-concept customization of text-to-image
  diffusion}. In \bibinfo{booktitle}{\emph{Proceedings of the IEEE/CVF
  Conference on Computer Vision and Pattern Recognition}}.
  \bibinfo{pages}{1931--1941}.
\newblock


\bibitem[\protect\citeauthoryear{Li, Li, and Hoi}{Li et~al\mbox{.}}{2024c}]%
        {blipdiff}
\bibfield{author}{\bibinfo{person}{Dongxu Li}, \bibinfo{person}{Junnan Li},
  {and} \bibinfo{person}{Steven Hoi}.} \bibinfo{year}{2024}\natexlab{c}.
\newblock \showarticletitle{Blip-diffusion: Pre-trained subject representation
  for controllable text-to-image generation and editing}.
\newblock \bibinfo{journal}{\emph{Advances in Neural Information Processing
  Systems}}  \bibinfo{volume}{36} (\bibinfo{year}{2024}).
\newblock


\bibitem[\protect\citeauthoryear{Li, Huang, Zhang, Hu, Liu, and Mao}{Li
  et~al\mbox{.}}{2024b}]%
        {li2024gradual}
\bibfield{author}{\bibinfo{person}{Hao Li}, \bibinfo{person}{Mengqi Huang},
  \bibinfo{person}{Lei Zhang}, \bibinfo{person}{Bo Hu}, \bibinfo{person}{Yi
  Liu}, {and} \bibinfo{person}{Zhendong Mao}.}
  \bibinfo{year}{2024}\natexlab{b}.
\newblock \showarticletitle{Gradual residuals alignment: a dual-stream
  framework for GAN inversion and image attribute editing}. In
  \bibinfo{booktitle}{\emph{Proceedings of the AAAI Conference on Artificial
  Intelligence}}, Vol.~\bibinfo{volume}{38}. \bibinfo{pages}{3064--3072}.
\newblock


\bibitem[\protect\citeauthoryear{Li, Hou, and Loy}{Li et~al\mbox{.}}{2024a}]%
        {w+}
\bibfield{author}{\bibinfo{person}{Xiaoming Li}, \bibinfo{person}{Xinyu Hou},
  {and} \bibinfo{person}{Chen~Change Loy}.} \bibinfo{year}{2024}\natexlab{a}.
\newblock \showarticletitle{When stylegan meets stable diffusion: a w+ adapter
  for personalized image generation}. In \bibinfo{booktitle}{\emph{Proceedings
  of the IEEE/CVF Conference on Computer Vision and Pattern Recognition}}.
  \bibinfo{pages}{2187--2196}.
\newblock


\bibitem[\protect\citeauthoryear{Liu, Song, and Chen}{Liu
  et~al\mbox{.}}{2023}]%
        {delving}
\bibfield{author}{\bibinfo{person}{Hongyu Liu}, \bibinfo{person}{Yibing Song},
  {and} \bibinfo{person}{Qifeng Chen}.} \bibinfo{year}{2023}\natexlab{}.
\newblock \showarticletitle{Delving stylegan inversion for image editing: A
  foundation latent space viewpoint}. In \bibinfo{booktitle}{\emph{Proceedings
  of the IEEE/CVF conference on computer vision and pattern recognition}}.
  \bibinfo{pages}{10072--10082}.
\newblock


\bibitem[\protect\citeauthoryear{Liu, Wang, Qian, Wang, and Rui}{Liu
  et~al\mbox{.}}{2024}]%
        {liu2024structure}
\bibfield{author}{\bibinfo{person}{Haipeng Liu}, \bibinfo{person}{Yang Wang},
  \bibinfo{person}{Biao Qian}, \bibinfo{person}{Meng Wang}, {and}
  \bibinfo{person}{Yong Rui}.} \bibinfo{year}{2024}\natexlab{}.
\newblock \showarticletitle{Structure matters: Tackling the semantic
  discrepancy in diffusion models for image inpainting}. In
  \bibinfo{booktitle}{\emph{Proceedings of the IEEE/CVF Conference on Computer
  Vision and Pattern Recognition}}. \bibinfo{pages}{8038--8047}.
\newblock


\bibitem[\protect\citeauthoryear{Liu, Wang, Wang, and Rui}{Liu
  et~al\mbox{.}}{2022}]%
        {liu2022delving}
\bibfield{author}{\bibinfo{person}{Haipeng Liu}, \bibinfo{person}{Yang Wang},
  \bibinfo{person}{Meng Wang}, {and} \bibinfo{person}{Yong Rui}.}
  \bibinfo{year}{2022}\natexlab{}.
\newblock \showarticletitle{Delving globally into texture and structure for
  image inpainting}. In \bibinfo{booktitle}{\emph{Proceedings of the 30th ACM
  International Conference on Multimedia}}. \bibinfo{pages}{1270--1278}.
\newblock


\bibitem[\protect\citeauthoryear{Long, Ye, Chen, Wang, Wang, and Yin}{Long
  et~al\mbox{.}}{2024}]%
        {diffusion1}
\bibfield{author}{\bibinfo{person}{Jing Long}, \bibinfo{person}{Guanhua Ye},
  \bibinfo{person}{Tong Chen}, \bibinfo{person}{Yang Wang},
  \bibinfo{person}{Meng Wang}, {and} \bibinfo{person}{Hongzhi Yin}.}
  \bibinfo{year}{2024}\natexlab{}.
\newblock \showarticletitle{Diffusion-based cloud-edge-device collaborative
  learning for next POI recommendations}. In
  \bibinfo{booktitle}{\emph{Proceedings of the 30th ACM SIGKDD Conference on
  Knowledge Discovery and Data Mining}}. \bibinfo{pages}{2026--2036}.
\newblock


\bibitem[\protect\citeauthoryear{Loshchilov}{Loshchilov}{2017}]%
        {adaw}
\bibfield{author}{\bibinfo{person}{I Loshchilov}.}
  \bibinfo{year}{2017}\natexlab{}.
\newblock \showarticletitle{Decoupled weight decay regularization}.
\newblock \bibinfo{journal}{\emph{arXiv preprint arXiv:1711.05101}}
  (\bibinfo{year}{2017}).
\newblock


\bibitem[\protect\citeauthoryear{Nichol, Dhariwal, Ramesh, Shyam, Mishkin,
  McGrew, Sutskever, and Chen}{Nichol et~al\mbox{.}}{2021}]%
        {GLIDE}
\bibfield{author}{\bibinfo{person}{Alex Nichol}, \bibinfo{person}{Prafulla
  Dhariwal}, \bibinfo{person}{Aditya Ramesh}, \bibinfo{person}{Pranav Shyam},
  \bibinfo{person}{Pamela Mishkin}, \bibinfo{person}{Bob McGrew},
  \bibinfo{person}{Ilya Sutskever}, {and} \bibinfo{person}{Mark Chen}.}
  \bibinfo{year}{2021}\natexlab{}.
\newblock \showarticletitle{Glide: Towards photorealistic image generation and
  editing with text-guided diffusion models}.
\newblock \bibinfo{journal}{\emph{arXiv preprint arXiv:2112.10741}}
  (\bibinfo{year}{2021}).
\newblock


\bibitem[\protect\citeauthoryear{Pehlivan, Dalva, and Dundar}{Pehlivan
  et~al\mbox{.}}{2023a}]%
        {styleres}
\bibfield{author}{\bibinfo{person}{Hamza Pehlivan}, \bibinfo{person}{Yusuf
  Dalva}, {and} \bibinfo{person}{Aysegul Dundar}.}
  \bibinfo{year}{2023}\natexlab{a}.
\newblock \showarticletitle{Styleres: Transforming the residuals for real image
  editing with stylegan}. In \bibinfo{booktitle}{\emph{Proceedings of the
  IEEE/CVF conference on computer vision and pattern recognition}}.
  \bibinfo{pages}{1828--1837}.
\newblock


\bibitem[\protect\citeauthoryear{Pehlivan, Dalva, and Dundar}{Pehlivan
  et~al\mbox{.}}{2023b}]%
        {pehlivan2023styleres}
\bibfield{author}{\bibinfo{person}{Hamza Pehlivan}, \bibinfo{person}{Yusuf
  Dalva}, {and} \bibinfo{person}{Aysegul Dundar}.}
  \bibinfo{year}{2023}\natexlab{b}.
\newblock \showarticletitle{Styleres: Transforming the residuals for real image
  editing with stylegan}. In \bibinfo{booktitle}{\emph{Proceedings of the
  IEEE/CVF conference on computer vision and pattern recognition}}.
  \bibinfo{pages}{1828--1837}.
\newblock


\bibitem[\protect\citeauthoryear{Qian, Wang, Hong, and Wang}{Qian
  et~al\mbox{.}}{2023a}]%
        {qian2023adaptive}
\bibfield{author}{\bibinfo{person}{Biao Qian}, \bibinfo{person}{Yang Wang},
  \bibinfo{person}{Richang Hong}, {and} \bibinfo{person}{Meng Wang}.}
  \bibinfo{year}{2023}\natexlab{a}.
\newblock \showarticletitle{Adaptive data-free quantization}. In
  \bibinfo{booktitle}{\emph{Proceedings of the IEEE/CVF Conference on Computer
  Vision and Pattern Recognition}}. \bibinfo{pages}{7960--7968}.
\newblock


\bibitem[\protect\citeauthoryear{Qian, Wang, Hong, and Wang}{Qian
  et~al\mbox{.}}{2023b}]%
        {qian2023rethinking}
\bibfield{author}{\bibinfo{person}{Biao Qian}, \bibinfo{person}{Yang Wang},
  \bibinfo{person}{Richang Hong}, {and} \bibinfo{person}{Meng Wang}.}
  \bibinfo{year}{2023}\natexlab{b}.
\newblock \showarticletitle{Rethinking data-free quantization as a zero-sum
  game}. In \bibinfo{booktitle}{\emph{Proceedings of the AAAI conference on
  artificial intelligence}}, Vol.~\bibinfo{volume}{37}.
  \bibinfo{pages}{9489--9497}.
\newblock


\bibitem[\protect\citeauthoryear{Radford, Kim, Hallacy, Ramesh, Goh, Agarwal,
  Sastry, Askell, Mishkin, Clark, et~al\mbox{.}}{Radford
  et~al\mbox{.}}{2021a}]%
        {clip}
\bibfield{author}{\bibinfo{person}{Alec Radford}, \bibinfo{person}{Jong~Wook
  Kim}, \bibinfo{person}{Chris Hallacy}, \bibinfo{person}{Aditya Ramesh},
  \bibinfo{person}{Gabriel Goh}, \bibinfo{person}{Sandhini Agarwal},
  \bibinfo{person}{Girish Sastry}, \bibinfo{person}{Amanda Askell},
  \bibinfo{person}{Pamela Mishkin}, \bibinfo{person}{Jack Clark},
  {et~al\mbox{.}}} \bibinfo{year}{2021}\natexlab{a}.
\newblock \showarticletitle{Learning transferable visual models from natural
  language supervision}. In \bibinfo{booktitle}{\emph{International conference
  on machine learning}}. PMLR, \bibinfo{pages}{8748--8763}.
\newblock


\bibitem[\protect\citeauthoryear{Radford, Kim, Hallacy, Ramesh, Goh, Agarwal,
  Sastry, Askell, Mishkin, Clark, et~al\mbox{.}}{Radford
  et~al\mbox{.}}{2021b}]%
        {clip-T}
\bibfield{author}{\bibinfo{person}{Alec Radford}, \bibinfo{person}{Jong~Wook
  Kim}, \bibinfo{person}{Chris Hallacy}, \bibinfo{person}{Aditya Ramesh},
  \bibinfo{person}{Gabriel Goh}, \bibinfo{person}{Sandhini Agarwal},
  \bibinfo{person}{Girish Sastry}, \bibinfo{person}{Amanda Askell},
  \bibinfo{person}{Pamela Mishkin}, \bibinfo{person}{Jack Clark},
  {et~al\mbox{.}}} \bibinfo{year}{2021}\natexlab{b}.
\newblock \showarticletitle{Learning transferable visual models from natural
  language supervision}. In \bibinfo{booktitle}{\emph{International conference
  on machine learning}}. PmLR, \bibinfo{pages}{8748--8763}.
\newblock


\bibitem[\protect\citeauthoryear{Ramesh, Dhariwal, Nichol, Chu, and
  Chen}{Ramesh et~al\mbox{.}}{2022}]%
        {DALL}
\bibfield{author}{\bibinfo{person}{Aditya Ramesh}, \bibinfo{person}{Prafulla
  Dhariwal}, \bibinfo{person}{Alex Nichol}, \bibinfo{person}{Casey Chu}, {and}
  \bibinfo{person}{Mark Chen}.} \bibinfo{year}{2022}\natexlab{}.
\newblock \showarticletitle{Hierarchical text-conditional image generation with
  clip latents}.
\newblock \bibinfo{journal}{\emph{arXiv preprint arXiv:2204.06125}}
  \bibinfo{volume}{1}, \bibinfo{number}{2} (\bibinfo{year}{2022}),
  \bibinfo{pages}{3}.
\newblock


\bibitem[\protect\citeauthoryear{Rombach, Blattmann, Lorenz, Esser, and
  Ommer}{Rombach et~al\mbox{.}}{2022}]%
        {SD}
\bibfield{author}{\bibinfo{person}{Robin Rombach}, \bibinfo{person}{Andreas
  Blattmann}, \bibinfo{person}{Dominik Lorenz}, \bibinfo{person}{Patrick
  Esser}, {and} \bibinfo{person}{Bj{\"o}rn Ommer}.}
  \bibinfo{year}{2022}\natexlab{}.
\newblock \showarticletitle{High-resolution image synthesis with latent
  diffusion models}. In \bibinfo{booktitle}{\emph{Proceedings of the IEEE/CVF
  conference on computer vision and pattern recognition}}.
  \bibinfo{pages}{10684--10695}.
\newblock


\bibitem[\protect\citeauthoryear{Ronneberger, Fischer, and Brox}{Ronneberger
  et~al\mbox{.}}{2015}]%
        {unet}
\bibfield{author}{\bibinfo{person}{Olaf Ronneberger}, \bibinfo{person}{Philipp
  Fischer}, {and} \bibinfo{person}{Thomas Brox}.}
  \bibinfo{year}{2015}\natexlab{}.
\newblock \showarticletitle{U-net: Convolutional networks for biomedical image
  segmentation}. In \bibinfo{booktitle}{\emph{Medical image computing and
  computer-assisted intervention--MICCAI 2015: 18th international conference,
  Munich, Germany, October 5-9, 2015, proceedings, part III 18}}. Springer,
  \bibinfo{pages}{234--241}.
\newblock


\bibitem[\protect\citeauthoryear{Ruiz, Li, Jampani, Pritch, Rubinstein, and
  Aberman}{Ruiz et~al\mbox{.}}{2023}]%
        {dreambooth}
\bibfield{author}{\bibinfo{person}{Nataniel Ruiz}, \bibinfo{person}{Yuanzhen
  Li}, \bibinfo{person}{Varun Jampani}, \bibinfo{person}{Yael Pritch},
  \bibinfo{person}{Michael Rubinstein}, {and} \bibinfo{person}{Kfir Aberman}.}
  \bibinfo{year}{2023}\natexlab{}.
\newblock \showarticletitle{Dreambooth: Fine tuning text-to-image diffusion
  models for subject-driven generation}. In
  \bibinfo{booktitle}{\emph{Proceedings of the IEEE/CVF conference on computer
  vision and pattern recognition}}. \bibinfo{pages}{22500--22510}.
\newblock


\bibitem[\protect\citeauthoryear{Saharia, Chan, Saxena, Li, Whang, Denton,
  Ghasemipour, Gontijo~Lopes, Karagol~Ayan, Salimans, et~al\mbox{.}}{Saharia
  et~al\mbox{.}}{2022}]%
        {Imagen}
\bibfield{author}{\bibinfo{person}{Chitwan Saharia}, \bibinfo{person}{William
  Chan}, \bibinfo{person}{Saurabh Saxena}, \bibinfo{person}{Lala Li},
  \bibinfo{person}{Jay Whang}, \bibinfo{person}{Emily~L Denton},
  \bibinfo{person}{Kamyar Ghasemipour}, \bibinfo{person}{Raphael
  Gontijo~Lopes}, \bibinfo{person}{Burcu Karagol~Ayan}, \bibinfo{person}{Tim
  Salimans}, {et~al\mbox{.}}} \bibinfo{year}{2022}\natexlab{}.
\newblock \showarticletitle{Photorealistic text-to-image diffusion models with
  deep language understanding}.
\newblock \bibinfo{journal}{\emph{Advances in neural information processing
  systems}}  \bibinfo{volume}{35} (\bibinfo{year}{2022}),
  \bibinfo{pages}{36479--36494}.
\newblock


\bibitem[\protect\citeauthoryear{Song, Meng, and Ermon}{Song
  et~al\mbox{.}}{2020}]%
        {ddim}
\bibfield{author}{\bibinfo{person}{Jiaming Song}, \bibinfo{person}{Chenlin
  Meng}, {and} \bibinfo{person}{Stefano Ermon}.}
  \bibinfo{year}{2020}\natexlab{}.
\newblock \showarticletitle{Denoising diffusion implicit models}.
\newblock \bibinfo{journal}{\emph{arXiv preprint arXiv:2010.02502}}
  (\bibinfo{year}{2020}).
\newblock


\bibitem[\protect\citeauthoryear{Tov, Alaluf, Nitzan, Patashnik, and
  Cohen-Or}{Tov et~al\mbox{.}}{2021}]%
        {e4e}
\bibfield{author}{\bibinfo{person}{Omer Tov}, \bibinfo{person}{Yuval Alaluf},
  \bibinfo{person}{Yotam Nitzan}, \bibinfo{person}{Or Patashnik}, {and}
  \bibinfo{person}{Daniel Cohen-Or}.} \bibinfo{year}{2021}\natexlab{}.
\newblock \showarticletitle{Designing an encoder for stylegan image
  manipulation}.
\newblock \bibinfo{journal}{\emph{ACM Transactions on Graphics (TOG)}}
  \bibinfo{volume}{40}, \bibinfo{number}{4} (\bibinfo{year}{2021}),
  \bibinfo{pages}{1--14}.
\newblock


\bibitem[\protect\citeauthoryear{Wang, Zhang, Fan, Wang, and Chen}{Wang
  et~al\mbox{.}}{2022}]%
        {hfgi}
\bibfield{author}{\bibinfo{person}{Tengfei Wang}, \bibinfo{person}{Yong Zhang},
  \bibinfo{person}{Yanbo Fan}, \bibinfo{person}{Jue Wang}, {and}
  \bibinfo{person}{Qifeng Chen}.} \bibinfo{year}{2022}\natexlab{}.
\newblock \showarticletitle{High-fidelity gan inversion for image attribute
  editing}. In \bibinfo{booktitle}{\emph{Proceedings of the IEEE/CVF conference
  on computer vision and pattern recognition}}. \bibinfo{pages}{11379--11388}.
\newblock


\bibitem[\protect\citeauthoryear{Wang, Qian, Liu, Rui, and Wang}{Wang
  et~al\mbox{.}}{2024}]%
        {wang2024unpacking}
\bibfield{author}{\bibinfo{person}{Yang Wang}, \bibinfo{person}{Biao Qian},
  \bibinfo{person}{Haipeng Liu}, \bibinfo{person}{Yong Rui}, {and}
  \bibinfo{person}{Meng Wang}.} \bibinfo{year}{2024}\natexlab{}.
\newblock \showarticletitle{Unpacking the gap box against data-free knowledge
  distillation}.
\newblock \bibinfo{journal}{\emph{IEEE Transactions on Pattern Analysis and
  Machine Intelligence}} (\bibinfo{year}{2024}).
\newblock


\bibitem[\protect\citeauthoryear{Xiao, Yin, Freeman, Durand, and Han}{Xiao
  et~al\mbox{.}}{2024}]%
        {fastcomposer}
\bibfield{author}{\bibinfo{person}{Guangxuan Xiao}, \bibinfo{person}{Tianwei
  Yin}, \bibinfo{person}{William~T Freeman}, \bibinfo{person}{Fr{\'e}do
  Durand}, {and} \bibinfo{person}{Song Han}.} \bibinfo{year}{2024}\natexlab{}.
\newblock \showarticletitle{Fastcomposer: Tuning-free multi-subject image
  generation with localized attention}.
\newblock \bibinfo{journal}{\emph{International Journal of Computer Vision}}
  (\bibinfo{year}{2024}), \bibinfo{pages}{1--20}.
\newblock


\bibitem[\protect\citeauthoryear{Xue, Song, Guo, Liu, Zong, Liu, and Luo}{Xue
  et~al\mbox{.}}{2024}]%
        {RAPHAEL}
\bibfield{author}{\bibinfo{person}{Zeyue Xue}, \bibinfo{person}{Guanglu Song},
  \bibinfo{person}{Qiushan Guo}, \bibinfo{person}{Boxiao Liu},
  \bibinfo{person}{Zhuofan Zong}, \bibinfo{person}{Yu Liu}, {and}
  \bibinfo{person}{Ping Luo}.} \bibinfo{year}{2024}\natexlab{}.
\newblock \showarticletitle{Raphael: Text-to-image generation via large mixture
  of diffusion paths}.
\newblock \bibinfo{journal}{\emph{Advances in Neural Information Processing
  Systems}}  \bibinfo{volume}{36} (\bibinfo{year}{2024}).
\newblock


\bibitem[\protect\citeauthoryear{Ye, Zhang, Liu, Han, and Yang}{Ye
  et~al\mbox{.}}{2023}]%
        {ip}
\bibfield{author}{\bibinfo{person}{Hu Ye}, \bibinfo{person}{Jun Zhang},
  \bibinfo{person}{Sibo Liu}, \bibinfo{person}{Xiao Han}, {and}
  \bibinfo{person}{Wei Yang}.} \bibinfo{year}{2023}\natexlab{}.
\newblock \showarticletitle{Ip-adapter: Text compatible image prompt adapter
  for text-to-image diffusion models}.
\newblock \bibinfo{journal}{\emph{arXiv preprint arXiv:2308.06721}}
  (\bibinfo{year}{2023}).
\newblock


\bibitem[\protect\citeauthoryear{Zhang, Zhang, Li, and Qiao}{Zhang
  et~al\mbox{.}}{2016}]%
        {mtcnn}
\bibfield{author}{\bibinfo{person}{Kaipeng Zhang}, \bibinfo{person}{Zhanpeng
  Zhang}, \bibinfo{person}{Zhifeng Li}, {and} \bibinfo{person}{Yu Qiao}.}
  \bibinfo{year}{2016}\natexlab{}.
\newblock \showarticletitle{Joint face detection and alignment using multitask
  cascaded convolutional networks}.
\newblock \bibinfo{journal}{\emph{IEEE signal processing letters}}
  \bibinfo{volume}{23}, \bibinfo{number}{10} (\bibinfo{year}{2016}),
  \bibinfo{pages}{1499--1503}.
\newblock


\bibitem[\protect\citeauthoryear{Zhang, Isola, Efros, Shechtman, and
  Wang}{Zhang et~al\mbox{.}}{2018}]%
        {lpips}
\bibfield{author}{\bibinfo{person}{Richard Zhang}, \bibinfo{person}{Phillip
  Isola}, \bibinfo{person}{Alexei~A Efros}, \bibinfo{person}{Eli Shechtman},
  {and} \bibinfo{person}{Oliver Wang}.} \bibinfo{year}{2018}\natexlab{}.
\newblock \showarticletitle{The unreasonable effectiveness of deep features as
  a perceptual metric}. In \bibinfo{booktitle}{\emph{Proceedings of the IEEE
  conference on computer vision and pattern recognition}}.
  \bibinfo{pages}{586--595}.
\newblock


\end{thebibliography}



\begin{thebibliography}{0}


\ifx \showCODEN    \undefined \def \showCODEN     #1{\unskip}     \fi
\ifx \showDOI      \undefined \def \showDOI       #1{#1}\fi
\ifx \showISBNx    \undefined \def \showISBNx     #1{\unskip}     \fi
\ifx \showISBNxiii \undefined \def \showISBNxiii  #1{\unskip}     \fi
\ifx \showISSN     \undefined \def \showISSN      #1{\unskip}     \fi
\ifx \showLCCN     \undefined \def \showLCCN      #1{\unskip}     \fi
\ifx \shownote     \undefined \def \shownote      #1{#1}          \fi
\ifx \showarticletitle \undefined \def \showarticletitle #1{#1}   \fi
\ifx \showURL      \undefined \def \showURL       {\relax}        \fi
\providecommand\bibfield[2]{#2}
\providecommand\bibinfo[2]{#2}
\providecommand\natexlab[1]{#1}
\providecommand\showeprint[2][]{arXiv:#2}

\end{thebibliography}



\begin{thebibliography}{34}


\ifx \showCODEN    \undefined \def \showCODEN     #1{\unskip}     \fi
\ifx \showDOI      \undefined \def \showDOI       #1{#1}\fi
\ifx \showISBNx    \undefined \def \showISBNx     #1{\unskip}     \fi
\ifx \showISBNxiii \undefined \def \showISBNxiii  #1{\unskip}     \fi
\ifx \showISSN     \undefined \def \showISSN      #1{\unskip}     \fi
\ifx \showLCCN     \undefined \def \showLCCN      #1{\unskip}     \fi
\ifx \shownote     \undefined \def \shownote      #1{#1}          \fi
\ifx \showarticletitle \undefined \def \showarticletitle #1{#1}   \fi
\ifx \showURL      \undefined \def \showURL       {\relax}        \fi
\providecommand\bibfield[2]{#2}
\providecommand\bibinfo[2]{#2}
\providecommand\natexlab[1]{#1}
\providecommand\showeprint[2][]{arXiv:#2}

\bibitem[Balaji et~al\mbox{.}(2022)]%
        {eDiff}
\bibfield{author}{\bibinfo{person}{Yogesh Balaji}, \bibinfo{person}{Seungjun
  Nah}, \bibinfo{person}{Xun Huang}, \bibinfo{person}{Arash Vahdat},
  \bibinfo{person}{Jiaming Song}, \bibinfo{person}{Qinsheng Zhang},
  \bibinfo{person}{Karsten Kreis}, \bibinfo{person}{Miika Aittala},
  \bibinfo{person}{Timo Aila}, \bibinfo{person}{Samuli Laine}, {et~al\mbox{.}}}
  \bibinfo{year}{2022}\natexlab{}.
\newblock \showarticletitle{ediff-i: Text-to-image diffusion models with an
  ensemble of expert denoisers}.
\newblock \bibinfo{journal}{\emph{arXiv preprint arXiv:2211.01324}}
  (\bibinfo{year}{2022}).
\newblock


\bibitem[Baykal et~al\mbox{.}(2023)]%
        {baykalclip}
\bibfield{author}{\bibinfo{person}{Ahmet~Canberk Baykal},
  \bibinfo{person}{Abdul~Basit Anees}, \bibinfo{person}{Duygu Ceylan},
  \bibinfo{person}{Erkut Erdem}, \bibinfo{person}{Aykut Erdem}, {and}
  \bibinfo{person}{Deniz Yuret}.} \bibinfo{year}{2023}\natexlab{}.
\newblock \showarticletitle{CLIP-guided StyleGAN Inversion for Text-driven Real
  Image Editing}.
\newblock \bibinfo{journal}{\emph{ACM Transactions on Graphics}}
  \bibinfo{volume}{42}, \bibinfo{number}{5} (\bibinfo{year}{2023}),
  \bibinfo{pages}{1--18}.
\newblock


\bibitem[Deng et~al\mbox{.}(2019)]%
        {arcface}
\bibfield{author}{\bibinfo{person}{Jiankang Deng}, \bibinfo{person}{Jia Guo},
  \bibinfo{person}{Niannan Xue}, {and} \bibinfo{person}{Stefanos Zafeiriou}.}
  \bibinfo{year}{2019}\natexlab{}.
\newblock \showarticletitle{Arcface: Additive angular margin loss for deep face
  recognition}. In \bibinfo{booktitle}{\emph{Proceedings of the IEEE/CVF
  conference on computer vision and pattern recognition}}.
  \bibinfo{pages}{4690--4699}.
\newblock


\bibitem[Gal et~al\mbox{.}(2022)]%
        {textualinversion}
\bibfield{author}{\bibinfo{person}{Rinon Gal}, \bibinfo{person}{Yuval Alaluf},
  \bibinfo{person}{Yuval Atzmon}, \bibinfo{person}{Or Patashnik},
  \bibinfo{person}{Amit~H Bermano}, \bibinfo{person}{Gal Chechik}, {and}
  \bibinfo{person}{Daniel Cohen-Or}.} \bibinfo{year}{2022}\natexlab{}.
\newblock \showarticletitle{An image is worth one word: Personalizing
  text-to-image generation using textual inversion}.
\newblock \bibinfo{journal}{\emph{arXiv preprint arXiv:2208.01618}}
  (\bibinfo{year}{2022}).
\newblock


\bibitem[Hao et~al\mbox{.}(2023)]%
        {vico}
\bibfield{author}{\bibinfo{person}{Shaozhe Hao}, \bibinfo{person}{Kai Han},
  \bibinfo{person}{Shihao Zhao}, {and} \bibinfo{person}{Kwan-Yee~K Wong}.}
  \bibinfo{year}{2023}\natexlab{}.
\newblock \showarticletitle{ViCo: Plug-and-play Visual Condition for
  Personalized Text-to-image Generation}.
\newblock \bibinfo{journal}{\emph{arXiv preprint arXiv:2306.00971}}
  (\bibinfo{year}{2023}).
\newblock


\bibitem[Ho et~al\mbox{.}(2020)]%
        {ddpm}
\bibfield{author}{\bibinfo{person}{Jonathan Ho}, \bibinfo{person}{Ajay Jain},
  {and} \bibinfo{person}{Pieter Abbeel}.} \bibinfo{year}{2020}\natexlab{}.
\newblock \showarticletitle{Denoising diffusion probabilistic models}.
\newblock \bibinfo{journal}{\emph{Advances in neural information processing
  systems}}  \bibinfo{volume}{33} (\bibinfo{year}{2020}),
  \bibinfo{pages}{6840--6851}.
\newblock


\bibitem[Karras(2019)]%
        {StyleGAN}
\bibfield{author}{\bibinfo{person}{Tero Karras}.}
  \bibinfo{year}{2019}\natexlab{}.
\newblock \showarticletitle{A Style-Based Generator Architecture for Generative
  Adversarial Networks}.
\newblock \bibinfo{journal}{\emph{arXiv preprint arXiv:1812.04948}}
  (\bibinfo{year}{2019}).
\newblock


\bibitem[Kumari et~al\mbox{.}(2023)]%
        {customediff}
\bibfield{author}{\bibinfo{person}{Nupur Kumari}, \bibinfo{person}{Bingliang
  Zhang}, \bibinfo{person}{Richard Zhang}, \bibinfo{person}{Eli Shechtman},
  {and} \bibinfo{person}{Jun-Yan Zhu}.} \bibinfo{year}{2023}\natexlab{}.
\newblock \showarticletitle{Multi-concept customization of text-to-image
  diffusion}. In \bibinfo{booktitle}{\emph{Proceedings of the IEEE/CVF
  Conference on Computer Vision and Pattern Recognition}}.
  \bibinfo{pages}{1931--1941}.
\newblock


\bibitem[Li et~al\mbox{.}(2024b)]%
        {blipdiff}
\bibfield{author}{\bibinfo{person}{Dongxu Li}, \bibinfo{person}{Junnan Li},
  {and} \bibinfo{person}{Steven Hoi}.} \bibinfo{year}{2024}\natexlab{b}.
\newblock \showarticletitle{Blip-diffusion: Pre-trained subject representation
  for controllable text-to-image generation and editing}.
\newblock \bibinfo{journal}{\emph{Advances in Neural Information Processing
  Systems}}  \bibinfo{volume}{36} (\bibinfo{year}{2024}).
\newblock


\bibitem[Li et~al\mbox{.}(2023)]%
        {blip2encoder}
\bibfield{author}{\bibinfo{person}{Junnan Li}, \bibinfo{person}{Dongxu Li},
  \bibinfo{person}{Silvio Savarese}, {and} \bibinfo{person}{Steven Hoi}.}
  \bibinfo{year}{2023}\natexlab{}.
\newblock \showarticletitle{Blip-2: Bootstrapping language-image pre-training
  with frozen image encoders and large language models}. In
  \bibinfo{booktitle}{\emph{International conference on machine learning}}.
  PMLR, \bibinfo{pages}{19730--19742}.
\newblock


\bibitem[Li et~al\mbox{.}(2024a)]%
        {w+}
\bibfield{author}{\bibinfo{person}{Xiaoming Li}, \bibinfo{person}{Xinyu Hou},
  {and} \bibinfo{person}{Chen~Change Loy}.} \bibinfo{year}{2024}\natexlab{a}.
\newblock \showarticletitle{When stylegan meets stable diffusion: a w+ adapter
  for personalized image generation}. In \bibinfo{booktitle}{\emph{Proceedings
  of the IEEE/CVF Conference on Computer Vision and Pattern Recognition}}.
  \bibinfo{pages}{2187--2196}.
\newblock


\bibitem[Liu et~al\mbox{.}(2023)]%
        {delving}
\bibfield{author}{\bibinfo{person}{Hongyu Liu}, \bibinfo{person}{Yibing Song},
  {and} \bibinfo{person}{Qifeng Chen}.} \bibinfo{year}{2023}\natexlab{}.
\newblock \showarticletitle{Delving stylegan inversion for image editing: A
  foundation latent space viewpoint}. In \bibinfo{booktitle}{\emph{Proceedings
  of the IEEE/CVF conference on computer vision and pattern recognition}}.
  \bibinfo{pages}{10072--10082}.
\newblock


\bibitem[Loshchilov(2017)]%
        {adaw}
\bibfield{author}{\bibinfo{person}{I Loshchilov}.}
  \bibinfo{year}{2017}\natexlab{}.
\newblock \showarticletitle{Decoupled weight decay regularization}.
\newblock \bibinfo{journal}{\emph{arXiv preprint arXiv:1711.05101}}
  (\bibinfo{year}{2017}).
\newblock


\bibitem[Nichol et~al\mbox{.}(2021)]%
        {GLIDE}
\bibfield{author}{\bibinfo{person}{Alex Nichol}, \bibinfo{person}{Prafulla
  Dhariwal}, \bibinfo{person}{Aditya Ramesh}, \bibinfo{person}{Pranav Shyam},
  \bibinfo{person}{Pamela Mishkin}, \bibinfo{person}{Bob McGrew},
  \bibinfo{person}{Ilya Sutskever}, {and} \bibinfo{person}{Mark Chen}.}
  \bibinfo{year}{2021}\natexlab{}.
\newblock \showarticletitle{Glide: Towards photorealistic image generation and
  editing with text-guided diffusion models}.
\newblock \bibinfo{journal}{\emph{arXiv preprint arXiv:2112.10741}}
  (\bibinfo{year}{2021}).
\newblock


\bibitem[Pehlivan et~al\mbox{.}(2023)]%
        {styleres}
\bibfield{author}{\bibinfo{person}{Hamza Pehlivan}, \bibinfo{person}{Yusuf
  Dalva}, {and} \bibinfo{person}{Aysegul Dundar}.}
  \bibinfo{year}{2023}\natexlab{}.
\newblock \showarticletitle{Styleres: Transforming the residuals for real image
  editing with stylegan}. In \bibinfo{booktitle}{\emph{Proceedings of the
  IEEE/CVF conference on computer vision and pattern recognition}}.
  \bibinfo{pages}{1828--1837}.
\newblock


\bibitem[Radford et~al\mbox{.}(2021a)]%
        {clip}
\bibfield{author}{\bibinfo{person}{Alec Radford}, \bibinfo{person}{Jong~Wook
  Kim}, \bibinfo{person}{Chris Hallacy}, \bibinfo{person}{Aditya Ramesh},
  \bibinfo{person}{Gabriel Goh}, \bibinfo{person}{Sandhini Agarwal},
  \bibinfo{person}{Girish Sastry}, \bibinfo{person}{Amanda Askell},
  \bibinfo{person}{Pamela Mishkin}, \bibinfo{person}{Jack Clark},
  {et~al\mbox{.}}} \bibinfo{year}{2021}\natexlab{a}.
\newblock \showarticletitle{Learning transferable visual models from natural
  language supervision}. In \bibinfo{booktitle}{\emph{International conference
  on machine learning}}. PMLR, \bibinfo{pages}{8748--8763}.
\newblock


\bibitem[Radford et~al\mbox{.}(2021b)]%
        {clip-T}
\bibfield{author}{\bibinfo{person}{Alec Radford}, \bibinfo{person}{Jong~Wook
  Kim}, \bibinfo{person}{Chris Hallacy}, \bibinfo{person}{Aditya Ramesh},
  \bibinfo{person}{Gabriel Goh}, \bibinfo{person}{Sandhini Agarwal},
  \bibinfo{person}{Girish Sastry}, \bibinfo{person}{Amanda Askell},
  \bibinfo{person}{Pamela Mishkin}, \bibinfo{person}{Jack Clark},
  {et~al\mbox{.}}} \bibinfo{year}{2021}\natexlab{b}.
\newblock \showarticletitle{Learning transferable visual models from natural
  language supervision}. In \bibinfo{booktitle}{\emph{International conference
  on machine learning}}. PmLR, \bibinfo{pages}{8748--8763}.
\newblock


\bibitem[Ramesh et~al\mbox{.}(2022)]%
        {DALL}
\bibfield{author}{\bibinfo{person}{Aditya Ramesh}, \bibinfo{person}{Prafulla
  Dhariwal}, \bibinfo{person}{Alex Nichol}, \bibinfo{person}{Casey Chu}, {and}
  \bibinfo{person}{Mark Chen}.} \bibinfo{year}{2022}\natexlab{}.
\newblock \showarticletitle{Hierarchical text-conditional image generation with
  clip latents}.
\newblock \bibinfo{journal}{\emph{arXiv preprint arXiv:2204.06125}}
  \bibinfo{volume}{1}, \bibinfo{number}{2} (\bibinfo{year}{2022}),
  \bibinfo{pages}{3}.
\newblock


\bibitem[Rathvon(2004)]%
        {pseudoword}
\bibfield{author}{\bibinfo{person}{Natalie Rathvon}.}
  \bibinfo{year}{2004}\natexlab{}.
\newblock \bibinfo{booktitle}{\emph{Early Reading Assessment. A Practitioner's
  Handbook.}}
\newblock \bibinfo{publisher}{ERIC}.
\newblock


\bibitem[Rombach et~al\mbox{.}(2022a)]%
        {SD}
\bibfield{author}{\bibinfo{person}{Robin Rombach}, \bibinfo{person}{Andreas
  Blattmann}, \bibinfo{person}{Dominik Lorenz}, \bibinfo{person}{Patrick
  Esser}, {and} \bibinfo{person}{Bj{\"o}rn Ommer}.}
  \bibinfo{year}{2022}\natexlab{a}.
\newblock \showarticletitle{High-resolution image synthesis with latent
  diffusion models}. In \bibinfo{booktitle}{\emph{Proceedings of the IEEE/CVF
  conference on computer vision and pattern recognition}}.
  \bibinfo{pages}{10684--10695}.
\newblock


\bibitem[Rombach et~al\mbox{.}(2022b)]%
        {ldm}
\bibfield{author}{\bibinfo{person}{Robin Rombach}, \bibinfo{person}{Andreas
  Blattmann}, \bibinfo{person}{Dominik Lorenz}, \bibinfo{person}{Patrick
  Esser}, {and} \bibinfo{person}{Bj{\"o}rn Ommer}.}
  \bibinfo{year}{2022}\natexlab{b}.
\newblock \showarticletitle{High-resolution image synthesis with latent
  diffusion models}. In \bibinfo{booktitle}{\emph{Proceedings of the IEEE/CVF
  conference on computer vision and pattern recognition}}.
  \bibinfo{pages}{10684--10695}.
\newblock


\bibitem[Ronneberger et~al\mbox{.}(2015)]%
        {unet}
\bibfield{author}{\bibinfo{person}{Olaf Ronneberger}, \bibinfo{person}{Philipp
  Fischer}, {and} \bibinfo{person}{Thomas Brox}.}
  \bibinfo{year}{2015}\natexlab{}.
\newblock \showarticletitle{U-net: Convolutional networks for biomedical image
  segmentation}. In \bibinfo{booktitle}{\emph{Medical image computing and
  computer-assisted intervention--MICCAI 2015: 18th international conference,
  Munich, Germany, October 5-9, 2015, proceedings, part III 18}}. Springer,
  \bibinfo{pages}{234--241}.
\newblock


\bibitem[Ruiz et~al\mbox{.}(2023)]%
        {dreambooth}
\bibfield{author}{\bibinfo{person}{Nataniel Ruiz}, \bibinfo{person}{Yuanzhen
  Li}, \bibinfo{person}{Varun Jampani}, \bibinfo{person}{Yael Pritch},
  \bibinfo{person}{Michael Rubinstein}, {and} \bibinfo{person}{Kfir Aberman}.}
  \bibinfo{year}{2023}\natexlab{}.
\newblock \showarticletitle{Dreambooth: Fine tuning text-to-image diffusion
  models for subject-driven generation}. In
  \bibinfo{booktitle}{\emph{Proceedings of the IEEE/CVF conference on computer
  vision and pattern recognition}}. \bibinfo{pages}{22500--22510}.
\newblock


\bibitem[Saharia et~al\mbox{.}(2022)]%
        {Imagen}
\bibfield{author}{\bibinfo{person}{Chitwan Saharia}, \bibinfo{person}{William
  Chan}, \bibinfo{person}{Saurabh Saxena}, \bibinfo{person}{Lala Li},
  \bibinfo{person}{Jay Whang}, \bibinfo{person}{Emily~L Denton},
  \bibinfo{person}{Kamyar Ghasemipour}, \bibinfo{person}{Raphael
  Gontijo~Lopes}, \bibinfo{person}{Burcu Karagol~Ayan}, \bibinfo{person}{Tim
  Salimans}, {et~al\mbox{.}}} \bibinfo{year}{2022}\natexlab{}.
\newblock \showarticletitle{Photorealistic text-to-image diffusion models with
  deep language understanding}.
\newblock \bibinfo{journal}{\emph{Advances in neural information processing
  systems}}  \bibinfo{volume}{35} (\bibinfo{year}{2022}),
  \bibinfo{pages}{36479--36494}.
\newblock


\bibitem[Schuhmann et~al\mbox{.}(2022)]%
        {clip-vencoder}
\bibfield{author}{\bibinfo{person}{Christoph Schuhmann},
  \bibinfo{person}{Romain Beaumont}, \bibinfo{person}{Richard Vencu},
  \bibinfo{person}{Cade~W Gordon}, \bibinfo{person}{Ross Wightman},
  \bibinfo{person}{Mehdi Cherti}, \bibinfo{person}{Theo Coombes},
  \bibinfo{person}{Aarush Katta}, \bibinfo{person}{Clayton Mullis},
  \bibinfo{person}{Mitchell Wortsman}, \bibinfo{person}{Patrick Schramowski},
  \bibinfo{person}{Srivatsa~R Kundurthy}, \bibinfo{person}{Katherine Crowson},
  \bibinfo{person}{Ludwig Schmidt}, \bibinfo{person}{Robert Kaczmarczyk}, {and}
  \bibinfo{person}{Jenia Jitsev}.} \bibinfo{year}{2022}\natexlab{}.
\newblock \showarticletitle{{LAION}-5B: An open large-scale dataset for
  training next generation image-text models}. In
  \bibinfo{booktitle}{\emph{Thirty-sixth Conference on Neural Information
  Processing Systems Datasets and Benchmarks Track}}.
\newblock
\urldef\tempurl%
\url{https://openreview.net/forum?id=M3Y74vmsMcY}
\showURL{%
\tempurl}


\bibitem[Song et~al\mbox{.}(2020)]%
        {ddim}
\bibfield{author}{\bibinfo{person}{Jiaming Song}, \bibinfo{person}{Chenlin
  Meng}, {and} \bibinfo{person}{Stefano Ermon}.}
  \bibinfo{year}{2020}\natexlab{}.
\newblock \showarticletitle{Denoising diffusion implicit models}.
\newblock \bibinfo{journal}{\emph{arXiv preprint arXiv:2010.02502}}
  (\bibinfo{year}{2020}).
\newblock


\bibitem[Tov et~al\mbox{.}(2021)]%
        {e4e}
\bibfield{author}{\bibinfo{person}{Omer Tov}, \bibinfo{person}{Yuval Alaluf},
  \bibinfo{person}{Yotam Nitzan}, \bibinfo{person}{Or Patashnik}, {and}
  \bibinfo{person}{Daniel Cohen-Or}.} \bibinfo{year}{2021}\natexlab{}.
\newblock \showarticletitle{Designing an encoder for stylegan image
  manipulation}.
\newblock \bibinfo{journal}{\emph{ACM Transactions on Graphics (TOG)}}
  \bibinfo{volume}{40}, \bibinfo{number}{4} (\bibinfo{year}{2021}),
  \bibinfo{pages}{1--14}.
\newblock


\bibitem[Wang et~al\mbox{.}(2022)]%
        {hfgi}
\bibfield{author}{\bibinfo{person}{Tengfei Wang}, \bibinfo{person}{Yong Zhang},
  \bibinfo{person}{Yanbo Fan}, \bibinfo{person}{Jue Wang}, {and}
  \bibinfo{person}{Qifeng Chen}.} \bibinfo{year}{2022}\natexlab{}.
\newblock \showarticletitle{High-fidelity gan inversion for image attribute
  editing}. In \bibinfo{booktitle}{\emph{Proceedings of the IEEE/CVF conference
  on computer vision and pattern recognition}}. \bibinfo{pages}{11379--11388}.
\newblock


\bibitem[Wei et~al\mbox{.}(2023)]%
        {elite}
\bibfield{author}{\bibinfo{person}{Yuxiang Wei}, \bibinfo{person}{Yabo Zhang},
  \bibinfo{person}{Zhilong Ji}, \bibinfo{person}{Jinfeng Bai},
  \bibinfo{person}{Lei Zhang}, {and} \bibinfo{person}{Wangmeng Zuo}.}
  \bibinfo{year}{2023}\natexlab{}.
\newblock \showarticletitle{Elite: Encoding visual concepts into textual
  embeddings for customized text-to-image generation}. In
  \bibinfo{booktitle}{\emph{Proceedings of the IEEE/CVF International
  Conference on Computer Vision}}. \bibinfo{pages}{15943--15953}.
\newblock


\bibitem[Xiao et~al\mbox{.}(2024)]%
        {fastcomposer}
\bibfield{author}{\bibinfo{person}{Guangxuan Xiao}, \bibinfo{person}{Tianwei
  Yin}, \bibinfo{person}{William~T Freeman}, \bibinfo{person}{Fr{\'e}do
  Durand}, {and} \bibinfo{person}{Song Han}.} \bibinfo{year}{2024}\natexlab{}.
\newblock \showarticletitle{Fastcomposer: Tuning-free multi-subject image
  generation with localized attention}.
\newblock \bibinfo{journal}{\emph{International Journal of Computer Vision}}
  (\bibinfo{year}{2024}), \bibinfo{pages}{1--20}.
\newblock


\bibitem[Xue et~al\mbox{.}(2024)]%
        {RAPHAEL}
\bibfield{author}{\bibinfo{person}{Zeyue Xue}, \bibinfo{person}{Guanglu Song},
  \bibinfo{person}{Qiushan Guo}, \bibinfo{person}{Boxiao Liu},
  \bibinfo{person}{Zhuofan Zong}, \bibinfo{person}{Yu Liu}, {and}
  \bibinfo{person}{Ping Luo}.} \bibinfo{year}{2024}\natexlab{}.
\newblock \showarticletitle{Raphael: Text-to-image generation via large mixture
  of diffusion paths}.
\newblock \bibinfo{journal}{\emph{Advances in Neural Information Processing
  Systems}}  \bibinfo{volume}{36} (\bibinfo{year}{2024}).
\newblock


\bibitem[Ye et~al\mbox{.}(2023)]%
        {ip}
\bibfield{author}{\bibinfo{person}{Hu Ye}, \bibinfo{person}{Jun Zhang},
  \bibinfo{person}{Sibo Liu}, \bibinfo{person}{Xiao Han}, {and}
  \bibinfo{person}{Wei Yang}.} \bibinfo{year}{2023}\natexlab{}.
\newblock \showarticletitle{Ip-adapter: Text compatible image prompt adapter
  for text-to-image diffusion models}.
\newblock \bibinfo{journal}{\emph{arXiv preprint arXiv:2308.06721}}
  (\bibinfo{year}{2023}).
\newblock


\bibitem[Zhang et~al\mbox{.}(2016)]%
        {mtcnn}
\bibfield{author}{\bibinfo{person}{Kaipeng Zhang}, \bibinfo{person}{Zhanpeng
  Zhang}, \bibinfo{person}{Zhifeng Li}, {and} \bibinfo{person}{Yu Qiao}.}
  \bibinfo{year}{2016}\natexlab{}.
\newblock \showarticletitle{Joint face detection and alignment using multitask
  cascaded convolutional networks}.
\newblock \bibinfo{journal}{\emph{IEEE signal processing letters}}
  \bibinfo{volume}{23}, \bibinfo{number}{10} (\bibinfo{year}{2016}),
  \bibinfo{pages}{1499--1503}.
\newblock


\bibitem[Zhang et~al\mbox{.}(2018)]%
        {lpips}
\bibfield{author}{\bibinfo{person}{Richard Zhang}, \bibinfo{person}{Phillip
  Isola}, \bibinfo{person}{Alexei~A Efros}, \bibinfo{person}{Eli Shechtman},
  {and} \bibinfo{person}{Oliver Wang}.} \bibinfo{year}{2018}\natexlab{}.
\newblock \showarticletitle{The unreasonable effectiveness of deep features as
  a perceptual metric}. In \bibinfo{booktitle}{\emph{Proceedings of the IEEE
  conference on computer vision and pattern recognition}}.
  \bibinfo{pages}{586--595}.
\newblock


\end{thebibliography}

\appendix

\end{document}


\renewcommand\footnotetextcopyrightpermission[1]{}
	\settopmatter{printacmref=false} 
	\title{Supplementary Material for \\ ``PIDiff: Image Customization for Personalized Identities with Diffusion Models''}
	
\clearpage
\setcounter{page}{1}

	
	\maketitle

\section{Overview}
Due to page limitation of the main body, the supplementary material offers further discussion on ablation study. The parameter settings and the training strategy play a crucial role in PIDiff. Therefore, we provide more analysis to validate the effectiveness of our method, which are summarized below:
\begin{itemize}
	\item Additional analysis of data augmentation, as mentioned in Sec.3.1 of the main body (Sec.\ref{sec1}).
	\item Additional analysis of training steps, as mentioned in Sec.3.1 of the main body (Sec.\ref{sec2}).
	\item Analysis of the number of images for training in style editing, as mentioned in Sec.3.1 of the main body (Sec.\ref{sec3}).
	\item Additionally, we provide the code to ensure the method can be reproduced (Sec.\ref{sec4}).
	\item We also provide the dataset to support the reproducibility of our experiments(Sec.\ref{sec5}).
\end{itemize}

\section{Additional Analysis of data augmentation}
\label{sec1}
\begin{figure}[H]
	\centering
	\includegraphics[width=0.45\textwidth]{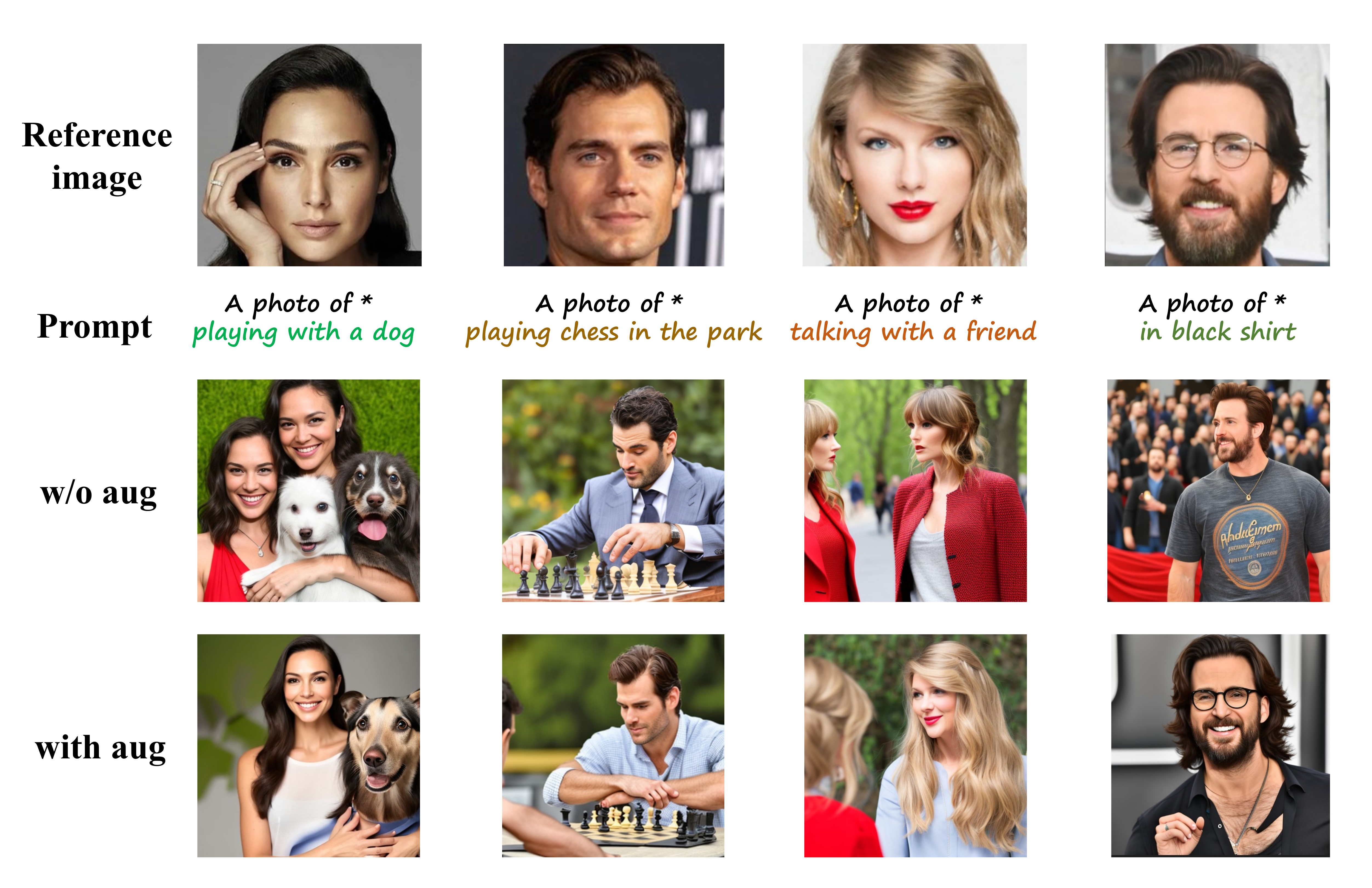}
	\caption{Qualitative results of using data augmentation. PIDiff can more accurately preserve identity features through data augmentation.}
	\label{aug}
\end{figure}

During training, we randomly add noise to ${w}_{+}$ vector. We drop text prompts with a probability of 0.3 and visual prompts with a probability of 0.5. First, the random noise added to ${w}_{+}$ vector effectively prevents overfitting of vector-image pairs and improve the diversity of generated images. We also discard the prompts with a high probability to avoid overfitting visual and text prompts and help the model customize to a specific identity. Fig. \ref{aug} shows the effect of our data augmentation. We also show the experiment results in table~\ref{tabaug}.

\begin{table}[H]
	\begin{center}
		\caption{Analysis of Data Augmentation}
		\label{tabaug}
		\begin{tabular}{c | c  c  c }
			\hline
			Methods & ID$\uparrow$ & LPIPS$\downarrow$ & CLIP-T$\uparrow$\\
			\hline
			w/o aug& 0.2744 & 0.6412 & 0.1649\\
			
			w/ aug & \textbf{0.3112} & \textbf{0.5936} & \textbf{0.1938} \\ 
			\hline
			
		\end{tabular}
	\end{center}
	
\end{table}

\section{Additional Analysis of Training Steps}
\label{sec2}

Different training strategies often require varying numbers of training steps. Previous methods require significantly different numbers of training steps. Therefore, we analyze how many steps are suitable for our model. 

\begin{figure}[H]
	\centering
	\includegraphics[width=\linewidth]{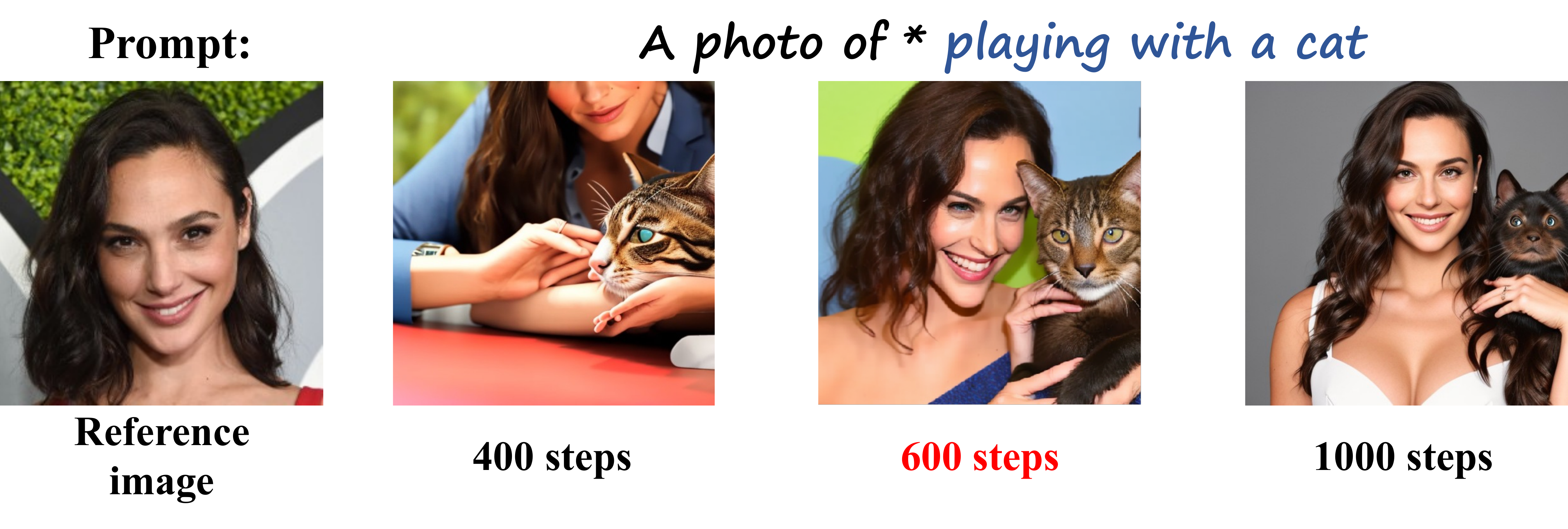}
	\caption{Qualitative results of using different training steps. Too few steps lead to identity loss. Too many steps cause overfitting, reducing text-image semantic consistency. }
	\label{steps}
\end{figure}

As shown in Table~\ref{stepstable}, we observe that when the number of training steps is fewer than 600, the model fails to integrate the specific identity into the pseudo-words, resulting in generated images that do not effectively capture the specific identity. Conversely, excessive training steps cause the model to overfit to the specific identity, leading to a decline in semantic alignment with text prompts (visual comparison is shown in Fig. \ref{steps}). 

To balance the ID and CLIP-T metrics, we select 600 training steps. This choice is based on the observation that our model achieves the best performance on CLIP-T while also generating high-quality images at 600 steps.

\begin{table}[H]
	\begin{center}
		\caption{Analysis of training steps for train}
		\label{stepstable}
		\begin{tabular}{c | c  c  c c }
			\hline
			number & 400 & 600 & 800  & 1000\\
			\hline
			ID$\uparrow$ & 0.2973 & 0.3112 & 0.3219 & \textbf{0.3390}\\
			
			CLIP-T$\uparrow$ & 0.1535 & \textbf{0.1938} & 0.1265 & 0.1144 \\ 
			\hline
			
		\end{tabular}
	\end{center}
\end{table}

\section{Additional Analysis of the Number of Images for Training in Style Editing}
\label{sec3}
One of the contributions of PIDiff is the implementation of style editing under a customized fine-tuning training strategy. PIDiff enables style editing by combining ${w}_{+}$ vectors from different images of a specific identity. However, if the desired style is absent in the images of the specific identity, an additional image is needed to provide the required style during training. Therefore, to ensure the preservation of the original identity features while learning the new style, we train the model using six images of the specific identity along with an additional face image containing the desired style. This method may lead to some degree of performance degradation,but as shown in Table~\ref{fusion}, our model still maintains strong performance.

\begin{table}[H]
	\begin{center}
		\caption{Analysis of Number of Images for Training (SI:Specific identity, ADD:An additional image of a new identity with a specific style)}
		\label{fusion}
		\begin{tabular}{c | c  c  c }
			\hline
			number & ID$\uparrow$ & LPIPS$\downarrow$ & CLIP-T$\uparrow$\\
			\hline
			6(SI)+1(ADD)& 0.30970 & 0.61467 & 0.18183\\
			
			6(SI) & \textbf{0.3112} & \textbf{0.5936} & \textbf{0.1938} \\ 
			\hline
			
		\end{tabular}
	\end{center}
	
\end{table}

\section{Code}
\label{sec4}
We implement the proposed PIDiff in pytorch framework under the running environment as: python 3.12.4, pytorch 2.2.2 and cuda 12.0. The codes are available in the package \textbf{code.zip}.

\section{Dataset}
\label{sec5}
We provide the dataset in the dataset folder. The images are organized into subfolders based on identity, ensuring clarity and ease of use for further analysis.

	\appendix